%% file: access.tex
\documentclass[conference,twocolumn]{IEEEtran}

\input{math_commands.tex}

\usepackage{url}
\usepackage{hyperref}
\usepackage{graphicx} % DO NOT CHANGE THIS
\usepackage{dirtytalk}

\usepackage[algo2e]{algorithm2e} 
\usepackage{algorithm}
\usepackage{algorithmic}

\usepackage{subfig}

\usepackage{array, multirow}
\usepackage{float}
\usepackage{tabularx}

\usepackage{amssymb}
\usepackage{makecell}

\usepackage{mathtools}

\newtheorem{definition}{Definition}
\ifCLASSINFOpdf
\else
\fi
\begin{document}

\title{Group Interventions on Deep Networks for Causal Discovery in Subsystems}

\author{
    \IEEEauthorblockN{Wasim Ahmad, Joachim Denzler, Maha Shadaydeh}
    \IEEEauthorblockA{ Computer Vision Group, Faculty of Mathematics and Computer Science, \\
    Friedrich Schiller University Jena, Germany
    \\wasim.ahmad, joachim.denzler, maha.shadaydeh}@uni-jena.de}

% \author{Wasim~Ahmad,~\IEEEmembership{}
%         Maha Shadaydeh,
%         and~Joachim~Denzler ~\IEEEmembership{}% <-this % stops a space
% \thanks{M. Shell was with the Department
% of Electrical and Computer Engineering, Georgia Institute of Technology, Atlanta,
% GA, 30332 USA e-mail: (see http://www.michaelshell.org/contact.html).}% <-this % stops a space
% \thanks{J. Doe and J. Doe are with Anonymous University.}% <-this % stops a space
% \thanks{Manuscript received April 19, 2005; revised August 26, 2015.}}

% \markboth{Journal of \LaTeX\ Class Files,~Vol.~14, No.~8, August~2015}%
% {Shell \MakeLowercase{\textit{et al.}}: Bare Demo of IEEEtran.cls for IEEE Journals}



\maketitle

\begin{abstract}
Causal discovery uncovers complex relationships between variables, enhancing predictions, decision-making, and insights into real-world systems, especially in nonlinear multivariate time series. However, most existing methods primarily focus on pairwise cause-effect relationships, overlooking interactions among groups of variables, i.e., subsystems and their collective causal influence. In this study, we introduce gCDMI, a novel multi-group causal discovery method that leverages group-level interventions on trained deep neural networks and employs model invariance testing to infer causal relationships. Our approach involves three key steps. First, we use deep learning to jointly model the structural relationships among groups of all time series. Second, we apply group-wise interventions to the trained model. Finally, we conduct model invariance testing to determine the presence of causal links among variable groups. We evaluate our method on simulated datasets, demonstrating its superior performance in identifying group-level causal relationships compared to existing methods. Additionally, we validate our approach on real-world datasets, including brain networks and climate ecosystems. Our results highlight that applying group-level interventions to deep learning models, combined with invariance testing, can effectively reveal complex causal structures, offering valuable insights for domains such as neuroscience and climate science.
\end{abstract}


% \fi
\IEEEpeerreviewmaketitle

-----------------------------
\section{Introduction}
\label{sec:introduction}
\begin{figure}[t]
\centering
\includegraphics[width=0.75\columnwidth]{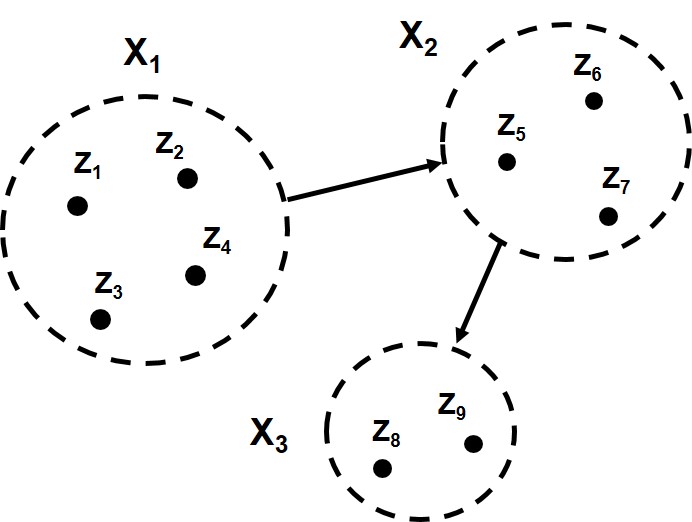}% 
\caption{Causal graph of a multivariate system of time series \(Z^N = \{Z_i\}_{i=1}^N\) organized into \(G = 3\) subsystems \(X^G = \{X_i\}_{i=1}^G\), where each subsystem \(X_i\) has dimension \(D_i\), which sums to the dimension of the whole system \(N = \sum_{i=1}^G D_i\).}
\label{fig:gcause}
\end{figure}

Causal discovery is the process of identifying cause-and-effect relationships from observational data. It plays a pivotal role in scientific research and real-world applications, aiding informed decision-making across fields like healthcare, economics, and policy. Most causal discovery methods focus on inferring causal relationships between individual variables. While effective in many cases, these approaches overlook the collective influence of groups of variables, where causal dynamics operate at the group level rather than at the level of individual variables, as illustrated in Figure \ref{fig:gcause}. It is particularly relevant when studying complex systems with interconnected components such as in climate-ecosystem \cite{molotoks2020comparing} and neuroscience \cite{faes2022new}, etc., allowing researchers to investigate how variables within distinct groups collectively contribute to observed outcomes or behaviors. Group-based causal analysis investigates causal relationships within specific groups of individuals, entities, or units. Grasping the causal interactions within these groups is crucial for modeling the emerging dynamics of these systems.

This work proposes a method to identify causal relationships within a multivariate time series system by analyzing it at the subsystem level. The system is assumed to be composed of multiple subsystems, each containing a specific subset of variables, referred to as a group of variables. Our approach consists of three key steps, as shown in Figure \ref{fig:schematic}: First, we learn the structure of temporal dependencies of all variables within groups (section \ref{structure-learning}); next, we apply group-wise interventions using in-distribution decorrelated data (section \ref{interventions}); and finally, we perform invariance testing between observational and interventional settings to infer causality (section \ref{model-invariance}). Building on our previous work \cite{ahmad2022causal}, which uses Knockoffs \cite{barber2015controlling, barber2019knockoff, barber2020robust} for pairwise causal discovery, we extend this framework to group-level causality—marking the first deep learning-based group causality method. Knockoffs are in-distribution variables that are decorrelated and generated independently of the output to eliminate any connection to it. The model invariance property refers to the invariant response of the model in the presence of causal predictors across different settings, i.e., interventions \cite{peters2016causal}. We employ DeepAR \cite{salinas2020deepar} for structural learning, capturing nonlinear interactions in multivariate time series data. To address deep networks' limitations with missing variables or out-of-distribution data, we adapt in-distribution, uncorrelated Knockoff interventions at the group level. This allows efficient testing of group-based causal hypotheses, significantly reducing the number of causal queries while enabling interpretable causal graphs. Our method also supports bidirectional causality, aiding the identification of feedback loops. Additionally, we integrate canonical correlation analysis (CCA) \cite{hardle2015canonical}, section \ref{multiset-canonical-correlation}, with pairwise causal discovery to manage high-dimensional groups, preserving correlation structures while reducing computational complexity. This enables pairwise causality methods to infer group-level causal relationships in a lower-dimensional space \cite{ashrafulla2013canonical, sato2010analyzing, zhuang2020technical}.
% -----------------------------------------------------
% -----------------------------------------------------

To demonstrate the robustness and practical utility of our approach against diverse networks, we evaluate its performance on synthetically generated and real-world time series datasets. First, we conduct experiments in two-group settings; where we compare our approach with two-group causality methods like Vanilla-PC \cite{janzing2009telling}, Trace method \cite{zscheischler2012testing}, and 2GVecCI \cite{wahl2023vector}. Since existing group causality methods typically handle only two groups, we employ multi-set canonical correlation (MCC) to reduce each group of variables into a single canonical representation. These canonical variables are then analyzed using pairwise causality methods such as VAR Granger Causality (VGC) \cite{ashrafulla2013canonical, lutkepohl2005new}, PCMCI \cite{runge2020discovering}, and CDMI \cite{ahmad2022causal}, allowing us to effectively compare their performance with our proposed approach.

Furthermore, we apply the methods to real-world scenarios, including climate-ecosystem interactions \cite{malhi2020climate, sefidmazgi2020causality, korell2020we} (Fluxnet \cite{pastorello2020Fluxnet2015}), the impact of the El Niño Southern Oscillation (ENSO 3.4) on sorrounding regions \cite{nowack2020causal}, and the influence of climate on tectonics, such as earth crust strain (Moxa \cite{kasburg2024trot}). Our approach is also validated on fMRI data (NetSim \cite{smith2011network}) to identify causal interactions in brain regions. Our method demonstrates improved performance across all datasets, with especially significant gains in simulated nonlinear data and brain data. In other datasets, the results are either better or comparable to alternative methods.
% -----------------------------------------------------------------

\section{Related Work}
\label{relatedwork}
\subsection{Pair-wise Causality Methods}

Causal discovery methods typically fall into constraint-based, score-based, and functional-based approaches. Constraint-based methods, like PC \cite{spirtes2000causation} and FCI \cite{spirtes2001anytime}, infer causal structures using conditional independence tests, with PCMCI \cite{runge2019detecting, runge2020discovering} extending this for time-series data. Score-based methods, such as Bayesian Networks, GES \cite{chickering2002optimal}, and LiNGAM \cite{shimizu2006linear}, optimize a scoring function to find the best-fitting causal structure. These methods perform well in pairwise causal discovery but struggle with high-dimensional datasets, where interactions occur within groups rather than between individual variables. Recently, deep learning methods have gained attention for their ability to model high-dimensional, nonlinear, and complex systems.

\subsection{Deep Learning-based Causal Discovery Methods}
These methods leverage deep neural networks to learn causal structures in a data-driven manner. In \cite{trifunov2022series}, the authors propose a method for estimating time series causal links under hidden confounding using sequential causal effect variational autoencoders (SCEVAEs). Causal Generative Neural Networks (CGNN) use generative models to infer causal relationships by learning the data distribution and comparing generated and observed data \cite{goudet2018learning}. The work of \cite{lachapelle2019gradient} extends variational autoencoders (VAEs) to learn latent representations while enforcing causal constraints, making it particularly useful for high-dimensional data. Temporal Causal Discovery using Recurrent Neural Networks (TCDF) applies RNNs with attention mechanisms to uncover nonlinear dependencies in sequential datasets \cite{nauta2019causal}.

% --------------------------------------------------------------
\begin{figure*}[t]
\centering
\includegraphics[width=0.85\textwidth]{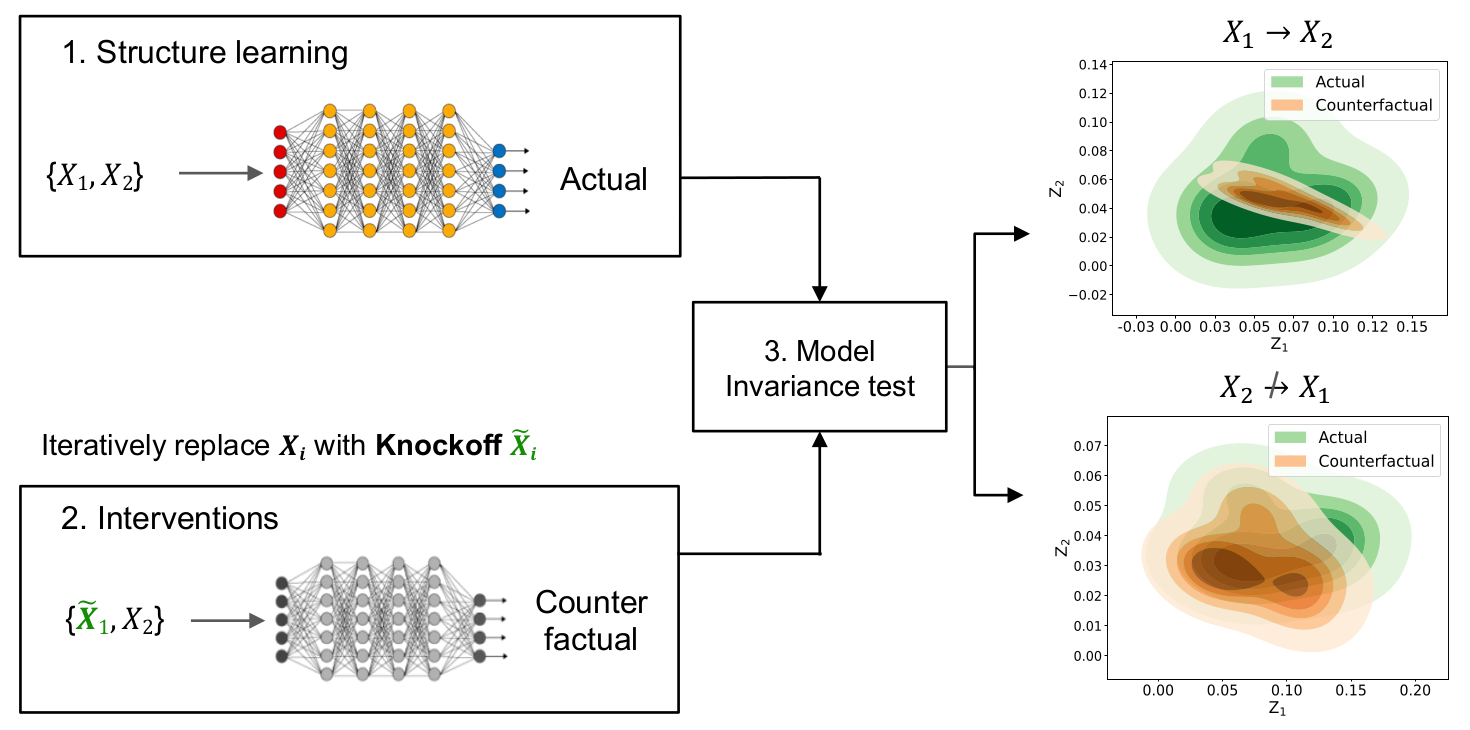}% 
\caption{A schematic of group causal discovery using deep networks for complex structure learning. Followed by interventions on the trained model. Causal relationship is assessed via model invariance testing by iteratively replacing \(X_i\) with a knockoff \(\tilde{X}_i\) and evaluating its effect on target group.} 
\label{fig:schematic}
\end{figure*}
% ---------------------------------------------------------------

% Group-based causal discovery
\subsection{Group Causality Methods}
Beyond pairwise causal interactions, group-level causal discovery is gaining attention. A common approach infers group causality by aggregating pairwise causal relationships, but this becomes computationally infeasible in large systems, leading to dense, uninterpretable graphs. For instance, in brain networks, a single region may contain numerous signals, making exhaustive pairwise analysis impractical and prone to spurious connections. An alternative is summarizing each group with a statistic, i.e., mean \cite{sato2010analyzing}, but this risks losing important inter-group dynamics. A more refined approach leverages dimensionality reduction, such as Canonical Correlation Analysis (CCA) \cite{hardle2015canonical}, which represents groups via canonical variables—linear combinations of within-group variables that maximize correlations with other groups. These variables then serve as inputs for causal discovery, enabling efficient group-level causal inference \cite{ashrafulla2013canonical, sato2010analyzing, zhuang2020technical}. Recent work \cite{Shadaydeh_2025} introduced a spectral causality approach to infer directional influence in fNIRS hyperscanning, achieving improved accuracy compared to time domain methods in motor tasks. Other methods for group causality include Vanilla-PC \cite{janzing2009telling}, Trace \cite{zscheischler2012testing}, and 2GVecCI \cite{wahl2023vector}, though they are limited to two-group cases. Additionally, VGC integrated with multi-set canonical correlation has been applied for group causal discovery in neuroscience \cite{ashrafulla2013canonical}.

% Our appraoch
Our work builds upon \cite{ahmad2022causal} by introducing a deep-learning-based framework for group-level causal discovery. Specifically, we leverage the potential of deep networks to model complex temporal dependencies within all variables representing the groups and apply group-based interventions aiming to find the causal impact of one group on another via invariance testing. Compared to prior methods, our approach is one of the first to apply deep learning for group-level causality, enabling better interpretability in complex systems such as climate-ecosystem interactions and neuroscience.
% ----------------------------------------------------------
\section{Method}
\label{methods}

This section presents our methodology for group causal discovery, starting with the problem definition and followed by a comprehensive breakdown of the steps taken to develop the solution, as summarised in Figure \ref{fig:schematic}.

\subsection{Problem definition}

We consider an \(N\)-variate time series, \(Z^N = \{Z_i\}_{i=1}^N\), which  are organized into \(G\) distinct groups or subsystems, \(X^G = \{X_i\}_{i=1}^G\), as illustrated in Figure \ref{fig:gcause}. Each group \(X_i\) is composed of \(D_i\) variables. The total number of variables across all groups is the sum of the dimensions of the individual groups: \(N = \sum_{i=1}^G D_i.\)

Each group represents the dynamics of a subsystem within an interconnected network. This grouping framework enables the study of interactions in subsystems, providing a way to model and analyze large, complex networks in a structured manner.

In this context, we define group causality as follows:

\begin{definition}[\textbf{Group Causality}]
\label{def:group_causality}
Let \( X^G = \{X_1, X_2, \dots, X_G\} \) represent the set of \( G \) groups of variables, where each group \( X_i = \{Z_i^1, Z_i^2, \dots, Z_i^{D_i}\} \) consists of \( D_i \) variables.  

We say that group \( X_i \) \textit{causes} group \( X_j \), denoted \( X_i \to X_j \), if there exists at least one variable \( Z_j^k \in X_j \) such that \( X_i \) causally influences \( Z_j^k \). Formally,  

\[
\exists Z_j^k \in X_j \text{ such that } X_i \to Z_j^k \quad \Rightarrow \quad X_i \to X_j.
\]
\end{definition}

Our method focuses on determining causal relationships in groups of variables, allowing for the identification of both uni-directional as well as bi-directional causal links:
\[
(X_i \to X_j) \land (X_j \to X_i) \quad \text{where } i \neq j.
\]

This flexibility enables the exploration of reciprocal interactions in complex systems, where causes and effects can switch over time. For example, in economics, the wage-price spiral occurs as rising wages increase demand, driving inflation, which then leads workers to demand higher wages, creating a continuous cycle.
% ----------------------------------- 
%            Assumptions 
% -----------------------------------
\subsection{Assumptions}
\begin{itemize}
    \item[-] \textbf{Stationarity}: We assume that the observed variables \( Z^N = \{Z_i\}_{i=1}^N \) form a stationary stochastic process, i.e., the mean \( \mu \) and variance \( \sigma^2 \) remain constant over time \( t \).

    \item[-] \textbf{Causal Sufficiency}: Furthermore, we assume that \(Z^N \) includes all relevant common causes in the system, i.e., causal sufficiency, ensuring the absence of hidden confounders. 

    \item[-] \textbf{Causal Coherence}: We also assume \textit{model invariance} under causal sufficiency, meaning the causal structure of the system remains consistent across different interventional environments.

    In other words, the causal relationships of the stochastic process \(Z^N\) stay stable even when specific variables or groups are modified. This invariance is crucial to our approach, as it helps identify robust and generalizable causal links across various conditions.
\end{itemize}

% ---------------------------------------------------------
% ---------------------------------------------------------
%                       Model Invariance 
% ---------------------------------------------------------
\subsection{Model Invariance}
\label{model-invariance}
In a causal framework, a model is considered invariant if, in the presence of its causal predictors, the distribution of its output residuals remains unchanged under interventional environments, as described in \cite{peters2016causal, meinshausen2016methods}. 
\[
\mathbb{P}(R_i \mid C, \mathcal{E}_i) = \mathbb{P}(\tilde{R}_j \mid C, \mathcal{E}_j)
\]
where \(R_i\), \(\tilde{R}_j\) represent residual distribution in observational and interventional environments \(\mathcal{E}_i\), \(\mathcal{E}_j\), and \(C\) is the set of causal predictors. More specifically, the response \(Z_i\) of the model (i.e., the target group of variables) remains invariant as long as none of the causally upstream group of variables are perturbed. This stability is assessed by examining the residual distribution, with any violation of the invariance property under interventions indicating a causal relationship.

\begin{definition}[Environment]
An \emph{environment} \(\mathcal{E}\) is a variant of the original system of multivariate time series \(\mathbf{Z} = (Z_1, Z_2, \ldots, Z_N)\), created by applying controlled interventions to one or more variables. Formally, \(\mathbf{Z}^{\mathcal{E}} = (Z_1^{\mathcal{E}}, Z_2^{\mathcal{E}}, \ldots, Z_N^{\mathcal{E}})\), where each \(Z_i^{\mathcal{E}}\) is generated to preserve key statistical or structural properties while isolating causal effects.
\end{definition}

% -------------------------------------------------------------

\subsection{Structure Learning}
\label{structure-learning}

To learn the structure of the entire system, we employ deep autoregressive models (DeepAR)\cite{salinas2020deepar}, recurrent neural networks (RNNs)-based approach to model the temporal dynamics of the system and generate a probabilistic forecast, e.g., the mean \(\mu\) and variance \(\sigma^2\) at each time step: 

\begin{equation}
    P(Z_i^{(t_0:T)} \mid Z_i^{(1:t_0 -1)}, Z_{-i}^{(1:t_0 -1)}; \Theta) \sim \mathcal{N} \\
\left( 
\begin{aligned}
    & \mu(Z_i, Z_{-i}; \Theta), \\
    & \sigma^2(Z_i, Z_{-i}; \Theta)
\end{aligned}
\right)
\end{equation}

which represents the future behavior of each variable \(Z_i\) over a forecast horizon \(t_0:T\), given its past values and the past values of all other variables (\(Z_{-i}\)) in the system. In this formulation, \(\Theta\) represents model parameters learned by the deep networks. This enables the model to capture the temporal dependencies of variables within the system. 
The mean (\(\mu\)) represents the central estimate of the system's state, while the variance (\(\sigma^2\)) quantifies the uncertainty in the predictions, which is critical for understanding the stochastic nature of the system.

Using the trained model, we compute the residuals for each variable \(Z_i\) over multiple forecast windows \(w_l\) (\(l = 1, \dots, L\)), denoted as \(R_i = \{e_{i1}, \dots, e_{il},\dots, e_{iL}\}\). The residuals for each variable \(Z_i\) are given by:

\[
e_{il}^{(t)} = \frac{1}{T} \sum_{t=1}^T \frac{|Z_i^{(t)} - \hat{Z}_i^{(t)}|}{Z_i^{(t)}},
\]

where \(\hat{Z}_i^{(t)}\) represents the predicted values of variable \(Z_i^{(t)}\), and \(T\) is the forecast horizon. The sequence \(R_i\) represents the forecast errors over multiple forecast windows, capturing the accuracy and uncertainty of the model's predictions for each group variable \(Z_i\).

% -------------------------------------------
%               Interventions 
% -------------------------------------------
\subsection{Group-based Interventions} 
\label{interventions}

As mentioned earlier, we have a multivariate system \(Z^N\), which is composed of \(G\) subsystems or groups, denoted as \(X^G\). For our causal query we define a target group \(X_j \in X^G\), while \(X^G_{-j}\) denotes all other groups. To assess the causal effect of a specific input group \(X_i \in X^G_{-j}
\) on target group \(X_j\), we replace \(X_i\) with its knockoffs \(\tilde{X}_i\), keeping the rest of the system unchanged:

\[
E\left[X_j \mid X^G_{-j}, do(X_i = \tilde{X}_i)\right]
\]

Here, \(do(X_i = \tilde{X}_i)\) represents an intervention where \(X_i\) is replaced with \(\tilde{X}_i\), a knockoff that is generated independent of the output. The term \(E[X_j]\) captures the expected behavior of the target group under this intervention, isolating the causal influence of \(X_i\) and facilitating the testing of causal relationships. The use of the \(do\)-operator \cite{pearl2000models} emphasizes that the intervention actively manipulates the input group of variables \(X_{i}\) using their knockoff counterpart \(\tilde{X}_{i}\), rather than passively observing their natural variations. The generation of knockoff variables \(\tilde{X}_i\) follows the framework proposed in \cite{barber2015controlling, barber2019knockoff, barber2020robust}. Knockoffs are constructed to serve as plausible but altered versions of the original variables. 

Specifically, the knockoff representation \(\tilde{X}_i\) preserves key statistical properties of the original group variables \(X_i\), such as their covariance structure, while ensuring they are in-distribution, uncorrelated to \(X_i\), and satisfy the exchangeability condition:  
\[
(X_i, \tilde{X}_i) \overset{d}{=} (\tilde{X}_i, X_i),
\]  
where \(\overset{d}{=}\) denotes equality in distribution. This property ensures that the original set of variables \(X_i\) and their knockoff counterparts \(\tilde{X}_i\) are indistinguishable in a probabilistic sense, enabling controlled interventions without introducing biases in causal analysis. The properties and diagnostics of knockoffs are illustrated in Appendix \ref{app:b}.

For our method, we generate knockoff copies \( \widetilde{Z}^N = \{\widetilde{Z}_1, \dots, \widetilde{Z}_N\} \) as interventional representations of all variables in the system \( Z^N \), which consists of \( G \) distinct groups \( X^G \). These knockoffs are used to systematically replace the original variables during the intervention process, allowing us to assess the collective causal impact of each input group \(X_{i}\) on the target group \(X_j\). The knockoff framework, as proposed in \cite{ahmad2021causal, ahmad2022causal}, has proven particularly effective for causal inference in time series data. These knockoffs act as controlled perturbations, enhancing causal analysis. The properties and effectiveness of knockoffs for interventions in causal inference are illustrated in Figure \ref{fig:knockoffs}, which highlights their exchangeability and statistical similarity to the original variables which ensures their suitability for accurate and unbiased causal analysis.

% <---------This part moved here from other section--------->
To test model invariance for causal links between groups, we apply the knockoff framework to generate interventional representations \(\tilde{X}_i\) for the candidate input group \(X_i\). For the target group \(X_j\), we determine its future values by conditioning on its past values and the interventional version of \(X_i\), while keeping other groups in \(X^G_{-j}\) unchanged.

% The conditional distribution of the counterfactuals for \(X_j\) is then expressed as:
% \[
% P(X_j^{(t+1:T)} \mid X_j^{(1:t)}, \tilde{X}_i^{(1:t)}, X_{-j \setminus i}^{(1:t)}; \Theta)
% \]
% where \(\tilde{X}_i^{(1:t)}\) represents the interventional representation of the specific group \(X_i\) (the only group being intervened on) while \(X_{-j \setminus i}^{(1:t)}\) shows the past values of all other groups in \(X_{-j}\), except \(X_i\), which remain unaltered. 

As a result of the intervention, we obtain the residuals for the counterfactual output, which are calculated as:  
\[
\widetilde{R}_j = \{\widetilde{e}_{j1}^{(t)}, \dots, \widetilde{e}_{jn}^{(t)}  \mid \widetilde{e}_j^{(t)} = \frac{|Z_j^{(t)} - \hat{\tilde{Z}}_j^{(t)}|}{Z_j^{(t)}}\}.
\]

The \textit{invariance property} states that if \( X_i \) does not causally affect \( X_j \), then the residual distributions of \( X_j \) remain unchanged before and after an intervention on \( X_i \). That is, the residuals \( R_j \) and \( \tilde{R}_j \) should be statistically indistinguishable:

\[
R_j \overset{d}{=} \widetilde{R}_j, \quad \text{if } X_i \text{ does not cause } X_j.
\]

\subsection{Causal Inference}
\label{subsec:causal-inference}

To infer the direction of the causal relationship, we applied various statistical hypothesis tests in our framework, details of which are given in the Appendix \ref{app:c}. 

\textbf{Hypothesis testing}: Based on our experiments, we chose the non-parametric Kolmogorov–Smirnov (KS) test \cite{smirnov1939estimation} to evaluate distribution shifts in the residuals of the variables of the target group \(X_j\) before and after interventions on the source group \(X_i\), without making any assumptions about the residual distributions. The significance testing allows us to assess whether a causal relationship exists. Specifically, we calculate the test statistic: 
 \[
   C_{ij} = \sqrt{\frac{pq}{p+q}} \sup |R_j - \tilde{R}_j|,
   \]
Here, \(p\) and \(q\) represent the number of samples in the distributions \(R_j\) and \(\tilde{R_j}\) respectively for variables in the taraget group \(X_j\), and \(\sup |R_j - \tilde{R_j}|\) measures the maximum absolute difference between these distributions. 

\textbf{Interpretation}: The KS test is performed at a significance level \(\alpha\) with the following hypotheses:

\begin{itemize}
    \item[-] Null Hypothesis (\(H_0\)): Assumes that \(X_i\) does not cause \(X_j\). This means that the residual distributions \(R_j\) and \(\tilde{R}_j\) are nearly identical, reflecting the invariance property under interventions in \(X_i\).
    \begin{itemize}
        \item Acceptance of \(H_0\): If the p-value from the test is greater than or equal to \(\alpha\), we fail to reject the null hypothesis, suggesting that \(X_i \nrightarrow X_j\).
    \end{itemize}
    \item[-] Alternative Hypothesis (\(H_1\)): States that \(X_i\) does cause \(X_j\). In this case, the intervention on \(X_i\) leads to a significant change in the residual distribution of \(X_j\), breaking the invariance property.
    \begin{itemize}
        \item Rejection of \(H_0\): If the p-value is less than \(\alpha\), we reject the null hypothesis in favor of the alternative, suggesting \(X_i \rightarrow X_j\).
    \end{itemize}
\end{itemize}

\textbf{Inference}: To determine causal direction, we test both \(X_i \rightarrow X_j\) and \(X_j \rightarrow X_i\) for all \( i \neq j \). This bidirectional testing helps resolve ambiguities in interactions between groups. A causal link is established if at least one variable in \(X_j\) is affected by \(X_i\), even if the influence is not uniform across all variables as stated in Definition \ref{def:group_causality}. That is, \(X_i \rightarrow X_j\) implies that interventions on \(X_i\) produce a measurable effect on some part of \(X_j\).  

With our causal framework, we can evaluate all possible relationship scenarios between \(X_i\) and \(X_j\), including \(X_i \rightarrow X_j\), \(X_j \rightarrow X_i\), \(X_i \leftrightarrow X_j\), or \(X_i \nleftrightarrow X_j\).

\begin{table*}[t]
    \centering
    \small
     \caption{F-scores for different methods across varying interaction densities on synthetically generated two-group data. The highest result for each interaction density is highlighted in bold.}
   \begin{tabular}{|l|c|c|c|c|c|c|c|c|c|c|}
\hline
\textbf{Method} & 
\multicolumn{9}{c|}{\textbf{Interaction density}}  \\ \cline{2-10}
                      &  \textbf{0.2} &  \textbf{0.3} &  \textbf{0.4} &  \textbf{0.5} &  \textbf{0.6} &  \textbf{0.7} &  \textbf{0.8} & \textbf{0.9} & \textbf{1.0} \\ \hline \hline
      \multirow{1}{*}{gCDMI} &  \textbf{1.00} &  0.67 &  \textbf{1.00} &  \textbf{1.00} & \textbf{1.00} &  \textbf{0.67} &  0.67 &  \textbf{1.00} &  \textbf{1.00} \\
      \hline
      \multirow{1}{*}{Trace} &  0.67 &  \textbf{1.00} &  0.67 &  0.67 &  0.67 &  \textbf{0.67} &  \textbf{1.00} &  0.67 &  0.67 \\
      \hline
    \multirow{1}{*}{2GVecCI} &  0.67 &  0.00 &  \textbf{1.00} &  0.33 &  0.67 &  0.33 &  \textbf{1.00} &  0.67 &  0.00 \\
    \hline
 \multirow{1}{*}{Vanilla-PC} &  0.67 &  0.67 &  0.67 &  0.33 &  0.67 & \textbf{0.67} &  0.33 &  0.33 &  0.00 \\
 \hline
\end{tabular}
    \label{tab:synresults}
\end{table*}

% ---------------------------------------------------------
%             Canonical Correlation Analysis
% ---------------------------------------------------------
\subsection{Multi-set Canonical Correlation}
\label{multiset-canonical-correlation}

We have a set of variables \( Z^N = \{Z_i\}_{i=1}^N \) organized into \( G \) groups, denoted \( X^G = \{X_i\}_{i=1}^G \). The goal is to identify canonical variables \( X_1^c, X_2^c, \dots, X_G^c \) for each group using canonical correlation analysis (CCA). Each canonical variable is a linear combination of the variables in its respective group \cite{hardle2015canonical}, and these combinations are optimized to maximize the correlations between canonical variables across groups. The process involves solving an eigenvalue problem to identify the linear combinations that best maximize the inter-group correlations.

The linear combinations are expressed as:

\[
X_i^c = A_i^T X_i
\]

where \( A_i \) is the vector of coefficients for group \( X_i \), and \( X_i \) is the vector of variables within the \( i_{th} \) group. 

CCA is formulated as a generalized eigenvalue problem, with the solution yielding the coefficients \( A_i \) that define the canonical variables for each group. The objective is to maximize the correlation between canonical variables \( X_1^c, X_2^c, \dots, X_G^c \) across all group pairs. This is achieved by selecting the coefficients \( A_1, A_2, \dots, A_G \) such that the correlations \( \text{corr}(X_i^c, X_j^c) \) for all distinct group pairs \( i \neq j \) are as high as possible.

The covariance matrix within each group is denoted \( \Sigma_{X_i X_i} \), while the cross-covariance between different groups is \( \Sigma_{X_i X_j} \). The generalized eigenvalue problem is given by:

\[
\Sigma_{X_i X_i}^{-1} \Sigma_{X_i X_j} \Sigma_{X_j X_j}^{-1} \Sigma_{X_i X_j}^T A = \lambda A
\]

where \( \lambda \) represents the eigenvalues corresponding to the squared canonical correlations, and \( A \) are the eigenvectors defining the optimal linear combinations. The eigenvectors corresponding to the largest eigenvalues are selected to maximize the inter-group correlations, ensuring the highest possible correlation between the canonical variables across groups. This aligns with the objective of CCA, which is to capture shared relationships between different groups of variables.
\section{Experiments}
\label{experiments}
To assess the performance of the methods, we conducted experiments using both synthetically generated time series and real-world datasets.

\subsection{Synthetic Data} For synthetic datasets, we use the following generic form of the structural causal models (SCMs) that provide a mathematical framework to represent causal relationships:
\[
Z_t^j \coloneqq \sum_i f_i(Z_{t-k_i}^i) + \eta_t^j, \quad i, j \in \{1, \dots, N\}, \quad 0 < k < t
\]
The system variables $Z_j$ have auto and cross-functional dependencies with a time delay of  $k$. The data model incorporates functional dependencies $f$, i.e., linear, trigonometric, and polynomial with varying interaction densities, and adds uncorrelated, normally distributed noise $\eta_t^j$. Interaction density refers to the proportion of actual causal edges among all possible edges, with higher density indicating more connections. Similarly, nonlinearity measures the fraction of nonlinear causal edges among all possible causal links, reflecting the complexity of interactions within the system. We evaluated the performance of these methods over varying data characteristics, i.e., interaction density, nonlinearity and groups variations.

\noindent \textbf{Increasing Interaction Density}: We first conducted experiments on causality analysis between two groups of time series and then extended our experiments to multiple groups. In our experiments, we explored different group dimensions for each edge density. We adjusted the deep network architecture according to each group dimension. Each method's decision for detecting causal edges between groups is categorized into one of three outcomes: correct (true positive), incorrect (false positive), or no detection (false negative).
To evaluate performance, we computed F-score values for each method across different group interaction densities, as the F-score accounts for class imbalance, which is particularly relevant in sparse graphs where causal links are few and non-causal relationships are prevalent.

In Table \ref{tab:synresults}, we have shown that our method (gCDMI) achieves better identification of the correct causal links with fewer incorrect detections in subsystems across all edge densities—ranging from sparse to dense causal graphs—compared to other group causality methods. Our method achieved the highest F-score in most cases, followed by the Trace method. Other methods showed lower F-scores due to false negatives, as it becomes challenging to distinguish cause from effect when the interaction between groups is complex. We also found that Vanilla-PC's performance degrades as edge density increases, while for other methods, increasing edge density does not have a clear impact on the outcome.

% % ----------------------------------------------------------------- 
% %        Synthetic: nonlinearity, interaction density, group size 
% % ------------------------------------------------------------
% \begin{figure*}[ht]
%     \centering
    
%     \subfloat{\subcaption{a} \includegraphics[width=0.3\textwidth]{figures/idense_fscore2.pdf}}
%      \subfloat{\subcaption{b} \includegraphics[width=0.3\textwidth]{figures/nonlin_fscore2.pdf}}
%     \subfloat{\subcaption{c} \includegraphics[width=0.3\textwidth]{figures/groups_fscore3.pdf}}

%     \caption{Performance of the methods over variation in interaction density, nonlinearity, and group size in the synthetically generated multi-groups dataset using structure causal models}
%     \label{fig:syn_multi_gcause}
% \end{figure*}
% % -------------------------------------------------------------

% --------------------------------------------------------- 
%          Synthetic multi-group experiments 
% ---------------------------------------------------------

% --------------------------------------------------------
%                 Interaction density
% --------------------------------------------------------
\begin{figure}
    \centering
    \includegraphics[width=\columnwidth]{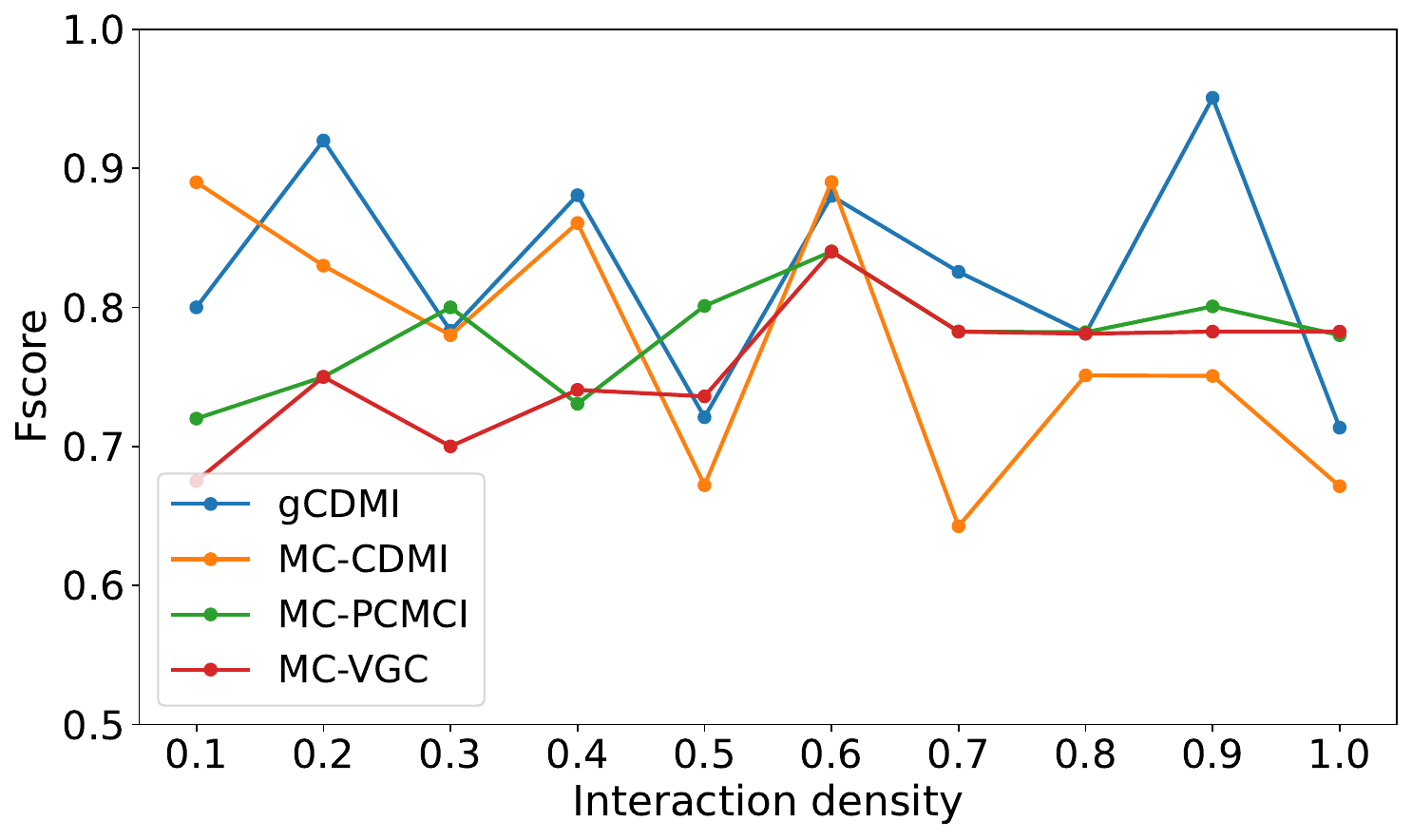}
    \caption{Performance of the methods over variation in interaction density in the synthetically generated multi-groups dataset using SCMs.}
    \label{fig:interaction}
\end{figure}
% ---------------------------------------------------------

Further, we carried out experiments on multi-group datasets, using four groups of two variables each and configuring the nonlinearity factor to 0.5. To assess performance, we examined scenarios with increasing interaction density, which introduced more causal edges among the groups and increased graph complexity. Here, we integrated CCA with pair-wise causality methods like PCMCI \cite{runge2019detecting}, VAR Granger Causality \cite{granger1969investigating, sims1980macroeconomics} and CDMI\cite{ahmad2022causal} to compare with our method gCDMI. Results for our method are shown in Figure \ref{fig:interaction} in comparison to MC-PCMCI, MC-VGC, and MC-CDMI. MC-VGC and MC-PCMCI exhibit a slight performance improvement as interaction densities increase. This is because both methods have a stronger tendency to positive discovery—regardless of whether they are true or false—which makes them more aligned with dense graphs. In contrast, gCDMI produces fluctuating results and does not exhibit a clear influence from increasing interaction density. However, despite these variations, it consistently outperforms other methods, maintaining a leading position across different scenarios. MC-CDMI demonstrates better initial performance; however, as the graph becomes increasingly dense, its effectiveness gradually declines. This deterioration suggests that while the method performs well in sparser networks, it struggles to maintain accuracy in highly interconnected structures. 

\begin{figure}[htp]
    \centering
    \includegraphics[width=\columnwidth]{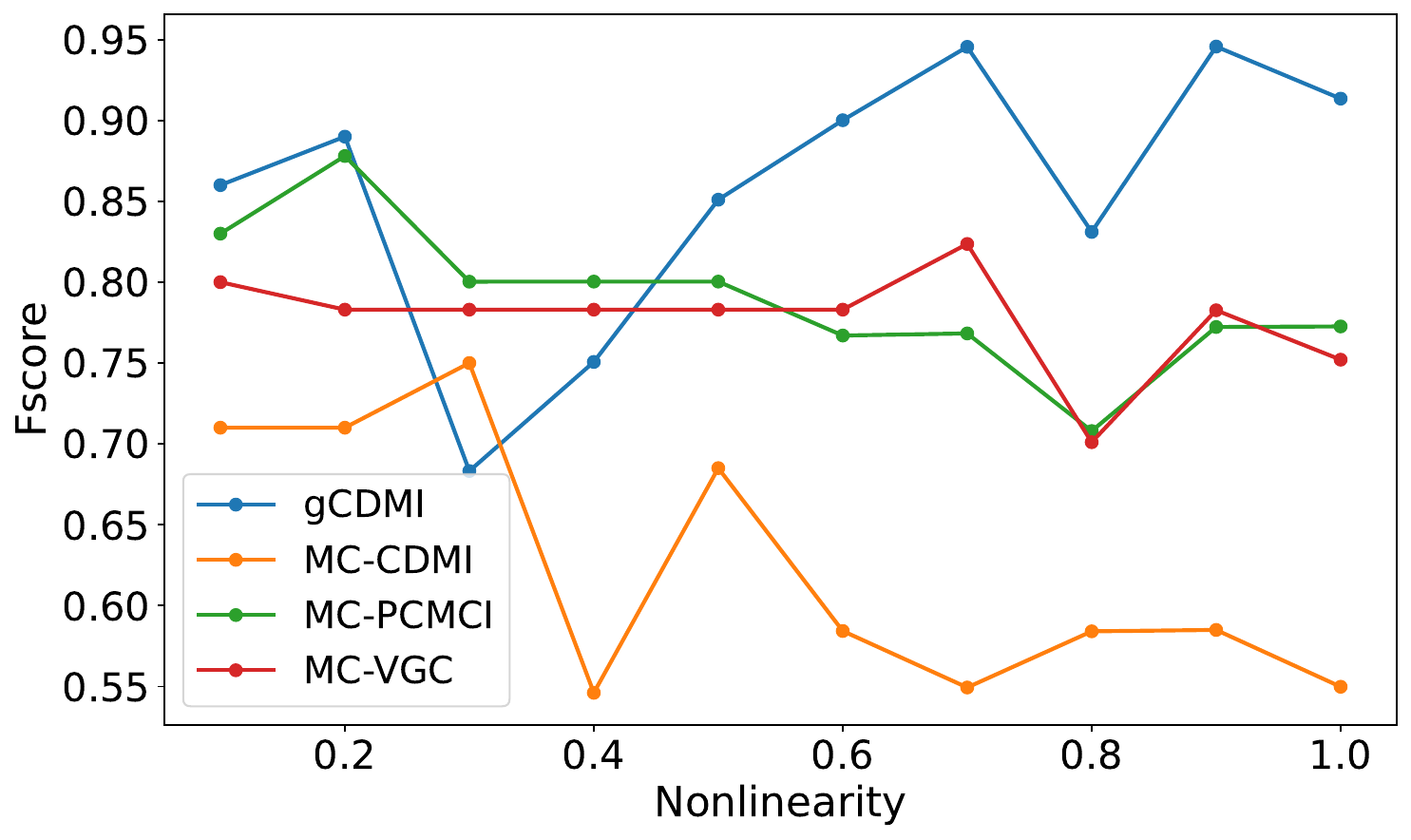}
    \caption{Performance of the methods over increasing nonlinearity in the synthetically generated multi-groups dataset.}
    \label{fig:nonlinearity}
\end{figure}

% ------------------------------------------------
%                 Increasing NOnlinearity
% ------------------------------------------------
\noindent  \textbf{Increasing Nonlinearity}: The experiment was conducted with four groups, each containing two variables, while the interaction density factor was set to 0.90. The performance is assessed by increasing the ratio of nonlinear edges in the graph. Figure \ref{fig:nonlinearity} illustrates the impact of increasing nonlinearity in the dataset. As nonlinearity rises, the robustness of different methods varies. gCDMI maintains high performance in most cases, indicating its strong suitability for capturing nonlinear relationships. In contrast, MC-CDMI experiences a significant drop in performance, highlighting its sensitivity to nonlinearity. MC-PCMCI and MC-VGC both demonstrate moderate performance with a slight decline and minimal fluctuations, suggesting a stable, however limited, ability to handle nonlinear dependencies. 

% -----------------Nonlinearity--------------------
% ---------------------------------------------------------

\begin{figure}
    \centering
    \includegraphics[width=\columnwidth]{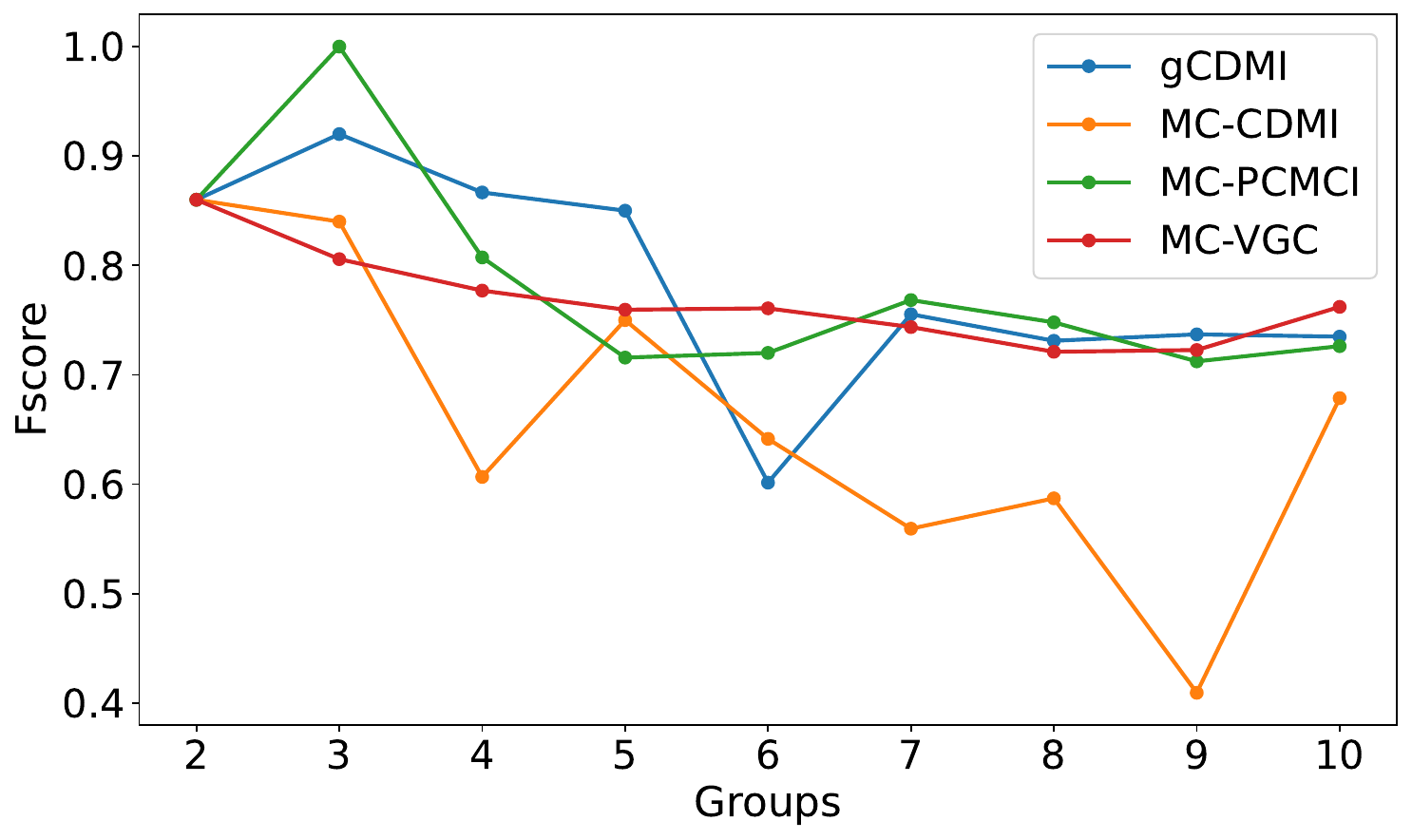}
    \caption{Performance of the methods over increasing number of groups in the synthetically generated multi-groups dataset.}
    \label{fig:groups}
\end{figure}

% ----------------------------------------------------
%                 GRoups variation
% ----------------------------------------------------
\noindent \textbf{Increasing Groups}: The experiment involved varying the number of groups, each containing two variables, with a nonlinearity factor of 0.5 and an interaction density of 0.9. Our objective was to evaluate each method's ability to uncover causal graphs as the number of groups increased. Figure \ref{fig:groups} evaluates the scalability of the methods by progressively increasing the groups. As the number of groups increases, all methods show a decline in performance. MC-VGC exhibits the slowest decline, indicating better scalability; however, it still performs worse than gCDMI and MC-PCMCI, which demonstrate moderate scalability but degrade more quickly. In contrast, MC-CDMI experiences a sharp drop in performance as group size increases. 

These results highlight the unique strengths and weaknesses of each method: gCDMI is highly effective in capturing nonlinear relationships, MC-PCMCI is the most scalable, and MC-VGC remains robust in denser graphs. However, MC-CDMI struggles in all high-complexity scenarios, particularly those involving increased nonlinearity, and high dimensionality from larger groups. Overall, this indicates that the characteristics of the datasets have a considerable impact on the performance of the methods with no clear all-scenario best method.
% --------------------------------------------------------
%                 GRoups variation
% --------------------------------------------------------
% ---------------------------------------------------------
% -----------------------------------------------------------------

\subsection{Real-world Data}
\noindent \textbf{Climate-Ecosystem}: 

Understanding how temperature and global radiation affect ecosystems is key to predicting their response to climate change and guiding environmental management. These interactions influence the carbon, water, and energy cycles, which are essential for climate modeling and policy-making. To investigate these dynamics, we conduct experiments using the \textit{Fluxnet2015} dataset~\cite{pastorello2020Fluxnet2015}, which provides high-quality measurements of carbon, water, and energy fluxes between the biosphere and atmosphere, collected via the eddy covariance method \cite{baldocchi2003assessing} across a network of global sites.
 
Data preparation is performed both at the individual site and network levels. For our experiments, we selected several measurement sites from the \textit{Fluxnet2015} dataset, including Hainich (DE-Hai), Monte Bondone (IT-MBo: Grasslands), Puechabon (FR-Pue: Evergreen Broadleaf Forests), and Tonzi Ranch (US-Ton: Woody Savannas). These sites were chosen for their ecological diversity, representing various ecosystem types across distinct geographic regions. The dataset includes climate and ecosystem time-series data such as global radiation (\(R_\text{g}\)), temperature (\(T\)), gross primary production (\(GPP\)), and ecosystem respiration (\(R_{\text{eco}}\)), recorded at multiple temporal resolutions (e.g., hourly, daily, weekly). 

We categorize these variables into two groups: the climate group (\(G_{\text{C}}\)), comprising temperature (\(T\)) and global radiation (\(R_\text{g}\)), and the ecosystem group (\(G_{\text{E}}\)), consisting of the ecosystem variables \(GPP\) and \(R_{\text{eco}}\). We opt for daily sampling frequency, which offers advantages in mitigating the influence of daily patterns that might obscure the underlying causal relationships in the data. 

The results in Table~\ref{tab:realdataresults} show the fractional occurrences of detected causal links in climate-ecosystem interactions across various sites and regimes from 2008 to 2010, as illustrated in Figure~\ref{fig:sitecomp} in Appendix~\ref{app:b}. To handle non-stationarity, we applied regime identification by clustering covariance matrices of multivariate time series, which helps reveal locally stable patterns. Regimes were identified using our previously proposed method~\cite{ahmad2024regime}, which improves the effectiveness of causality analysis, as demonstrated in Appendix~\ref{app:b}. Causal discovery was performed within each regime, and results were aggregated across all identified regimes.

\begin{figure}
    \centering
    \includegraphics[width=\columnwidth]{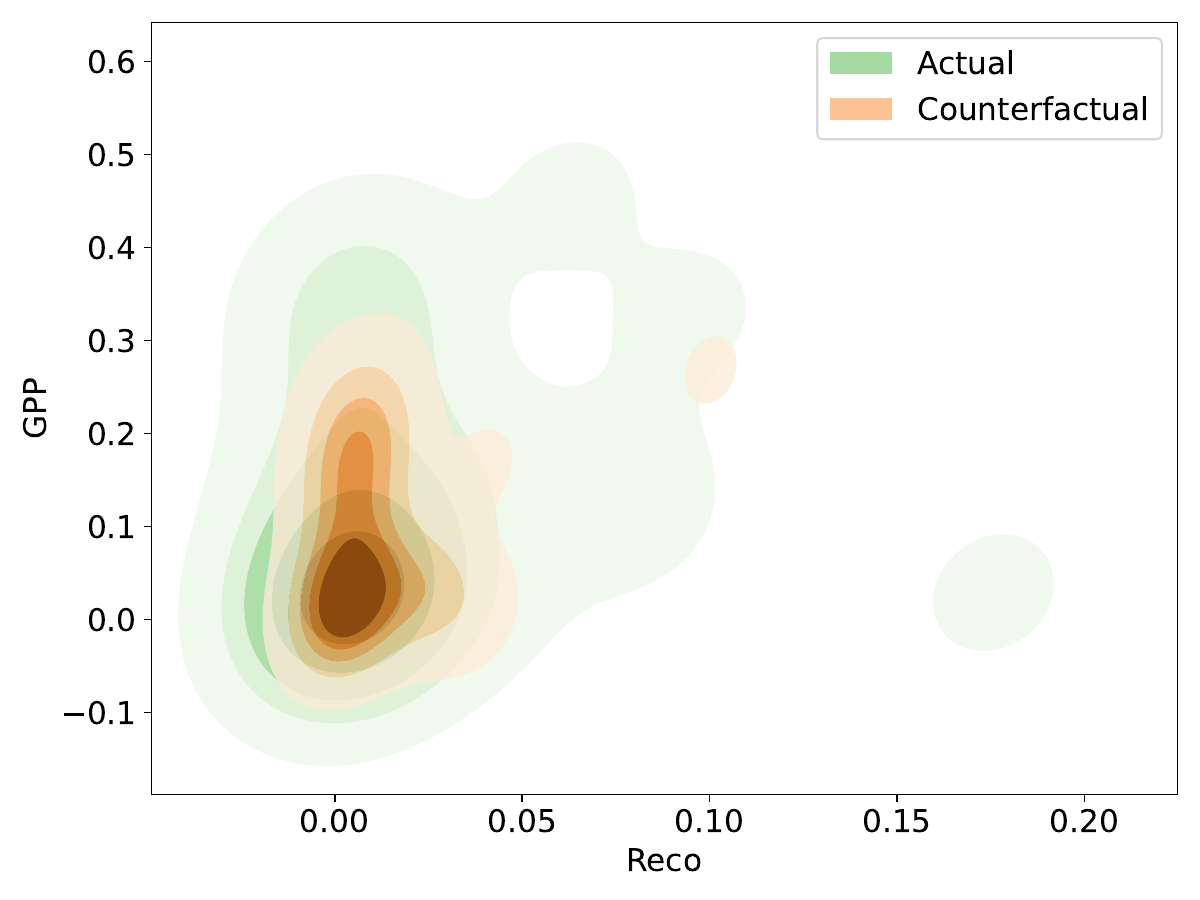}
    \caption{Distribution shift. Ecosystem group $G_{\text{E}} = \{GPP, R_{\text{eco}}\}$ in response to intervention on climate group $G_{\text{C}}= \{T, R_g\}$.}
    \label{fig:jointdistshift}
\end{figure}

The gCDMI method shows a notable prevalence of bidirectional causal links ($G_{\text{C}} \leftrightarrow G_{\text{E}}$), suggesting robust mutual interactions between climate and ecosystem variables, specifically in forest sites. In contrast, the 2GVecCI method effectively identifies causal links ($G_{\text{C}} \rightarrow G_{\text{E}}$) primarily in the DE-Hai site, but struggles to detect such links in other sites. The Trace method consistently highlights causal links ($G_{\text{C}} \rightarrow G_{\text{E}}$) across all sites, along with gCDMI for grasslands. Conversely, Vanilla-PC demonstrates limited success in identifying climate-ecosystem links, with notable detection only in the US-Ton site. As an illustration of climatic influence on the ecosystem (IT-MBo site) from one of our experiments, we show distribution shift in actual and counterfactual distributions in Figure \ref{fig:jointdistshift}.

\begin{table}
\centering
\small
\caption{Fractional occurrences of detected causal links in climate-ecosystem interactions across multiple sites over all regimes (2008–2010). Only gCDMI supports and reports bi-directional causal links, while other methods are limited to unidirectional detection.}
% \begin{adjustbox}{}
\begin{tabular}{llllll}
    \hline
     & & \multicolumn{4}{c}{Methods}\\
     \cline{3-6}
     \textbf{Sites} & \textbf{Links} & \textbf{gCDMI} & \textbf{2GVCI} & \textbf{Trace} & \textbf{V-PC} \\
    \hline
    \hline
    \multirow{2}{*}{\makecell{DE-Hai}} & $G_{\text{C}} \rightarrow G_{\text{E}}$ & 0.17 & 0.66 & 0.50 & 0.00 \\
    \cline{2-6}
    & $G_{\text{C}} \leftarrow G_{\text{E}}$  & 0.00 & 0.17 & 0.50 & 0.17 \\
    \cline{2-6}
    & $G_{\text{C}} \leftrightarrow G_{\text{E}}$  & 0.83 & $-$ & $-$ & $-$  \\
      \cline{2-6}
    & $G_{\text{C}} \nleftrightarrow G_{\text{E}}$  & 0.00 & 0.17 & 0.00 & 0.83 \\
    \hline
    \multirow{2}{*}{\makecell{IT-MBo}} & $G_{\text{C}} \rightarrow G_{\text{E}}$ & 0.66 & 0.34 & 0.83 &  0.00 \\
    \cline{2-6}
    & $G_{\text{C}} \leftarrow G_{\text{E}}$  & 0.17 & 0.00 & 0.17  & 0.00 \\
    \cline{2-6}
    & $G_{\text{C}} \leftrightarrow G_{\text{E}}$  & 0.17 & $-$ & $-$ & $-$  \\
    \cline{2-6}
    & $G_{\text{C}} \nleftrightarrow G_{\text{E}}$  & 0.00 & 0.66 & 0.00 & 1.00 \\
    \hline
    \multirow{2}{*}{\makecell{FR-Pue}} & $G_{\text{C}} \rightarrow G_{\text{E}}$ & 0.34 & 0.34 & 1.00 & 0.50 \\
    \cline{2-6}
    & $G_{\text{C}} \leftarrow G_{\text{E}}$  & 0.00 & 0.00 & 0.00 & 0.17 \\
     \cline{2-6}
    & $G_{\text{C}} \leftrightarrow G_{\text{E}}$  & 0.66 & $-$ & $-$ & $-$  \\
    \cline{2-6}
    & $G_{\text{C}} \nleftrightarrow G_{\text{E}}$  & 0.00 & 0.66 & 0.00 & 0.33  \\
    \hline
     \multirow{2}{*}{\makecell{US-Ton}} & $G_{\text{C}} \rightarrow G_{\text{E}}$ & 0.34 & 0.34 & 1.00 & 0.50 \\
    \cline{2-6}
    & $G_{\text{C}} \leftarrow G_{\text{E}}$ & 0.00 & 0.00 & 0.00 & 0.17\\
     \cline{2-6}
    & $G_{\text{C}} \leftrightarrow G_{\text{E}}$  & 0.66 & $-$ & $-$ & $-$  \\
    \cline{2-6}
    & $G_{\text{C}} \nleftrightarrow G_{\text{E}}$ & 0.00 & 0.66 & 0.00 & 0.33 \\
    \hline
    
\end{tabular}
% \end{adjustbox}
\label{tab:realdataresults}
\vspace{-1mm}%
\end{table}
% ---------------------------------------------------------------
\noindent \textbf{Tectonic-Climate Interactions}: We further examine how climatic and groundwater influence tectonic movements, providing insights into the complex, nonlinear interactions between Earth's crust and environmental conditions.
To achieve this, we utilize the highly sensitive laser strainmeter data from the Moxa Geodynamic Observatory (MGO) \cite{kasburg2024trot}. The strainmeters, installed in a low mountain overburden area, measure movements in the Earth's upper crust, but their readings are significantly influenced by local meteorological conditions. 

We employ group-causality methods to investigate causal interactions among the multivariate tectonic-climate variables listed in Table~\ref{tab:geoclimate}, aiming to uncover the nonlinear effects of climatic factors on strain observations. The climatic data, originally sampled at 10-second intervals, is downsampled to hourly resolution and normalized. As climatic factors may influence strain differently along the east-west and north-south axes, we perform separate analyses for each direction. This study uses three years of data (2014–2017), following the preprocessing pipeline described in~\cite{ahmad2024deep}.

To facilitate a more structured and interpretable causal analysis, we opted for group-level causal discovery, as opposed to the pairwise analysis used in~\cite{ahmad2024deep}. This approach better captures high-level interactions between related sets of variables and reduces the risk of spurious or fragmented causal links that may arise when treating variables in isolation. We organized the data into three distinct groups: tectonic (\(G_{\text{tec}}\))—comprising Strain\(_{ns}\) and Strain\(_{ew}\); climate (\(G_c\))—including temperature, global radiation, pressure, and wind speed; and groundwater wells (\(G_{\text{gw}}\))—containing GW\(_{mb}\) and GW\(_{west}\). We hypothesize that climate and groundwater variations (\(G_c, G_{\text{gw}}\)) causally influence tectonic activity (\(G_{\text{tec}}\)), a relationship we evaluate experimentally. 

To address non-stationarity, we used the regime identification method proposed in~\cite{ahmad2024regime}, which segments the data into locally stationary regimes. Following~\cite{ahmad2024deep}, we identified six regimes that capture distinct temporal dynamics, enabling more robust causal inference across varying system states.

% % -----------------------------------------------------------
% %          Boxplot: Fscore (Tectonic-climatic) 
% % -----------------------------------------------------------
% \begin{figure}[hbt]
% \centering
% \includegraphics[width=0.30
% \textwidth]{figures/causal_graph.pdf}% 
% \caption{The causal interaction in climate \(G_{c}\), tectonic \(G_{tec}\) and GW \(G_{gw}\) groups  (\(G_c\), \(G_{gw}\)) \(\rightarrow\) \(G_{tec}\).} 
% \label{fig:tectonic-climate-graph}
% \end{figure}
% % -----------------------------------------------------------
% % -----------------------------------------------------------
\begin{table}[h]
\centering
\caption{Time series of strainmeters, groundwater levels (GWLs), and meteorological data used in this study.}
\label{tab:geoclimate}
\begin{tabular}{|l|l|l|}
\hline
               & \textbf{Time series}                      & \textbf{Location}\\ 
\hline
\textbf{LSM}   & Strain$_{ns}$, Strain$_{ew}$              & Inside gallery   \\
\hline
\textbf{GWLs}  & GW$_{mb}$    & Outside gallery  \\
               & GW$_{west}$  & Inside gallery   \\
\hline
\textbf{Meteorological} & T (temperature),  & Weather station  \\
\textbf{data}          & P (barometric pressure),   &          \\ 
                       & R$_g$ (global radiation),  &          \\
                       & Wind                       &          \\           
\hline
\end{tabular}
\end{table}
% --------------------------------------------------------

% % ------------------------------------------------------
% \begin{figure*}
% \centering
% \includegraphics[width=\textwidth]{figures/jointdistshift3.png}% 
% \caption{Distribution shift of  \textbf{a}.  ecosystem group $G_{\text{E}} = \{GPP, R_{\text{eco}}\}$ in response to intervention on climate group  $G_{\text{C}}= \{T, R_g\} $.  \textbf{b}. British Columbia temperature in response to intervention on ENSO temperature data. \textbf{c}. Brain network $N_2$ in fMRI time series after group-level intervention on network $N_1$.} 
% \label{fig:jointdistshift}
% \end{figure*}
% % ------------------------------------------------------
% % --------------------------------------------------------
% \begin{figure*}[ht]
%     \centering
    
%     \subfloat{\subcaption{a} \includegraphics[width=0.32\textwidth]{figures/Group-1---Reco_2d.pdf}}
%      \subfloat{\subcaption{b} \includegraphics[width=0.32\textwidth]{figures/Group-1---BCT_2_2d.pdf}}
%     \subfloat{\subcaption{c} \includegraphics[width=0.3\textwidth]{figures/dist_synthetic.jpeg}}

%     \caption{Distribution shift of  \textbf{a}.  ecosystem group $G_{\text{E}} = \{GPP, R_{\text{eco}}\}$ in response to intervention on climate group  $G_{\text{C}}= \{T, R_g\} $.  \textbf{b}. British Columbia temperature in response to intervention on ENSO temperature data. \textbf{c}. Brain network $N_2$ in fMRI time series after group-level intervention on network $N_1$.}
%    \label{fig:jointdistshift}
% \end{figure*}
% ---------------------------------------------------------
% ---------------------------------------------------------

\begin{figure}[hbt]
\centering
\includegraphics[width=0.49
\textwidth]{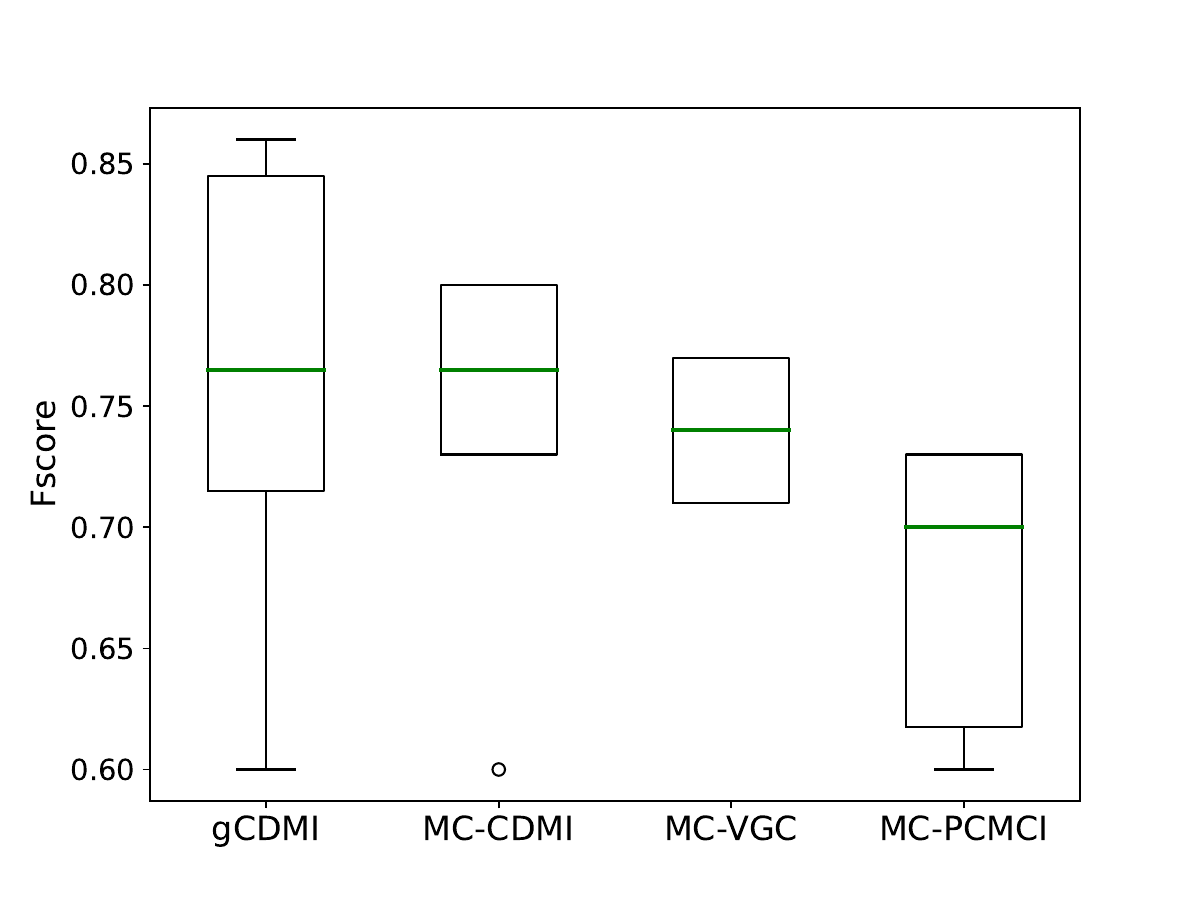}% 
\caption{F-score of the methods over all regimes in the tectonic-climate dataset.} 
\label{fig:tectonic-climate}
\end{figure}

The results, illustrated in Figure~\ref{fig:tectonic-climate}, present F-scores across all six regimes, providing insights into the performance and stability of the methods under varying conditions. While gCDMI achieves the highest F-score, it fluctuates more than MC-CDMI across all regimes. The analysis reveals that gCDMI along with MC-CDMI yield better results compared to MC-PCMCI and MC-VGC on average, reinforcing the expected causal influence of climate (\(G_c\)) and groundwater (\(G_{gw}\)) on tectonic activities (\(G_{tec}\)). This aligns with the hypothesis that environmental and subsurface dynamics play a crucial role in driving movements within the Earth's crust. In contrast, while MC-VGC also shows relatively stable performance compared to MC-PCMCI, it does not achieve the same level of F-score as the gCDMI and MC-CDMI methods. MC-PCMCI, on the other hand, exhibits greater variability in performance across regimes, indicating sensitivity to regime-specific characteristics. Overall, the findings underscore the effectiveness of incorporating group-level causal interventions (gCDMI) and canonical variable approaches (MC-CDMI) in identifying and reinforcing causal relationships, particularly in complex systems characterized by regime transitions.
% -----------------------------------------------------------
% -----------------------------------------------------------
%          Boxplot: Fscore (Tectonic-climatic) 
% -----------------------------------------------------------
% -----------------------------------------------------------

\textbf{ENSO effects on British Columbia}: We performed experiments on the El Niño–Southern Oscillation (ENSO) dataset, focusing on surface temperatures in the tropical Pacific ENSO region and British Columbia (BCT) from 1948 to 2021. ENSO refers to a recurring climate pattern involving changes in sea surface temperatures and atmospheric pressure in the equatorial Pacific Ocean. Its warm phase, known as El Niño, is associated with significant ocean warming that alters global atmospheric circulation. This disruption is known to influence climate variability in regions such as British Columbia \cite{taylor1998effect}.

For our experiments, we followed data preprocessing steps such as deseasonalization, smoothing, and aggregation, adapting methodologies from \cite{runge2019detecting, wahl2023vector}. The experimental results, presented in Table \ref{tab:enso}, show that both our method and 2GVecCI identified a significant influence of ENSO on BCT across different grid scales, achieving 66\% true positives and 34\% false negatives. In comparison, the Trace method detected the causal link ENSO \(\rightarrow\) BCT with 50\% true positives and 50\% false positives. Vanilla-PC, however, failed to infer any causal direction across all grid scales, likely due to difficulties in distinguishing data patterns between the two regions. 
% Figure \ref{fig:jointdistshift} (b) illustrates the causal influence of ENSO on BCT for one of our tests. It compares the BCT distribution with and without group-level intervention on the ENSO time series variables. The observed shift in BCT's counterfactual distribution serves as evidence of ENSO’s impact.

% -------------------------------------------------- 
%             Table: ENSO experiments 
% --------------------------------------------------
\begin{table}[hbt!]
\centering
%\Large
\caption{Proportion of detected causal links from ENSO region surface temperature to British Columbia across grid scales. Values indicate the detection ratio for each method.}

% \begin{adjustbox}{}
\begin{tabular}{|l|l|l|l|l|}
    \hline
    \textbf{Inference} & \textbf{gCDMI} & \textbf{Trace} & \textbf{2GVecCI} & \textbf{Vanilla-PC} \\
    \hline
    ENSO $\rightarrow$ BCT & 0.66 & 0.50 & 0.66 & 0.00\\
    \hline
    ENSO $\leftarrow$ BCT & 0.00 & 0.50 & 0.00 & 0.00\\
    \hline
    ENSO $\nleftrightarrow$ BCT & 0.34 & 0.00 & 0.34 & 1.00\\
    \hline
    
\end{tabular}
% \end{adjustbox}
\label{tab:enso}
\end{table}
% ------------------------------------------------

% % ---------------------------------------------- 
% %                Dyadic: fNIRS 
% % ----------------------------------------------
% \begin{figure*}[ht]
%     \centering
%     % Top-left figure
%       \subfloat{\includegraphics[width=0.44\textwidth]{figures/mean_acc_barplot.pdf}}
%     \subfloat{\includegraphics[width=0.44\textwidth]{figures/mean_acc_barplot.pdf}} \\
%     % Top-right figure
%     % Bottom-left figure
%     \subfloat{\includegraphics[width=0.44\textwidth]{figures/mean_acc_barplot.pdf}}
%     % Bottom-right figure
%     \subfloat{\includegraphics[width=0.44\textwidth]{figures/mean_acc_barplot.pdf}}

%     \caption{Accuracy of the methods for estimating dyadic interaction.}
%     \label{fig:2x2}
% \end{figure*}
% % ----------------------------------------------

\noindent \textbf{Brain Regions Interactions}: This analysis helps improve our understanding of brain network interactions, aiding in the accurate identification of causal relationships between brain regions. It is crucial for advancing brain connectivity research and developing better approaches for diagnosing and treating neurological disorders. We analyzed simulated fMRI time series data (NetSim) \cite{smith2011network} to detect interactions between brain network nodes and determine their directionality. These fMRI time series simulations were generated using the dynamic causal modeling (DCM) framework, which models neural interactions and hemodynamic responses to produce realistic fMRI data \cite{friston2003dynamic}. The generated dataset provides a ground truth connectivity graph for a variety of network topologies, which we use to evaluate our methods. Here, we conducted experiments on a number of simulated subjects from \textit{S}2 topology in the dataset, which contains 10 nodes clustered into 2 groups. It has 10 min fMRI sessions for each subject with 3 sec sampling rate, ﬁnal added noise of 1\%, and haemodynamic response function (HRF) variability of ±0.5 which provides time series of 200 data points. 

The fraction of method outcomes in terms of true positives, false positives, and false negatives is provided in Table \ref{tab:fmri}. Our approach produced 56\% true positives, 17\% false positives, and 21\% bidirectional links, which is an overall better performance compared to other applied group causality methods. Results from Trace were somehow comparable to that of gCDMI. While 2GVecCI and Vanilla-PC yielded a high percentage of false negatives, probably because of difficulty in separating cause from effect in complex brain signals. 

% The causal influence of brain network $N_1$ on $N_2$ for one subject in \textit{S}2 topology from simulated fMRI time series is demonstrated in Fig. \ref{fig:jointdistshift} (c).
% ---------------------------------------------- 
%           Table: FMRI experiments 
% ----------------------------------------------
\begin{table}[hbt!]
\centering
%\large
\caption{The table shows the fraction of decisions made by group causality methods for identifying brain network connections, with the ground truth \(N_1 \rightarrow N_2\).}
% \begin{adjustbox}{}
\begin{tabular}{|l|l|l|l|l|}
    \hline
    \textbf{Inference} & \textbf{gCDMI} & \textbf{Trace} & \textbf{2GVCI} & \textbf{V-PC} \\
    \hline
    $N_1 \rightarrow N_2$ &  0.56 & 0.50 &  0.33 &  0.06\\
    \hline
    $N_1 \leftarrow N_2$ & 0.17 & 0.39 & 0.17 & 0.16\\
    \hline
    $N_1 \leftrightarrow N_2$ & 0.21 & - & - & - \\
    \hline
    $N_1 \nleftrightarrow N_2$ & 0.06 & 0.11 & 0.50 & 0.78\\
    \hline
    
\end{tabular}
% \end{adjustbox}
\label{tab:fmri}
\end{table} 
% -------------------------------------------------
%                multi-group NetSIm
% -------------------------------------------------

\begin{figure}[hbt]
\centering
\includegraphics[width=0.5\textwidth]{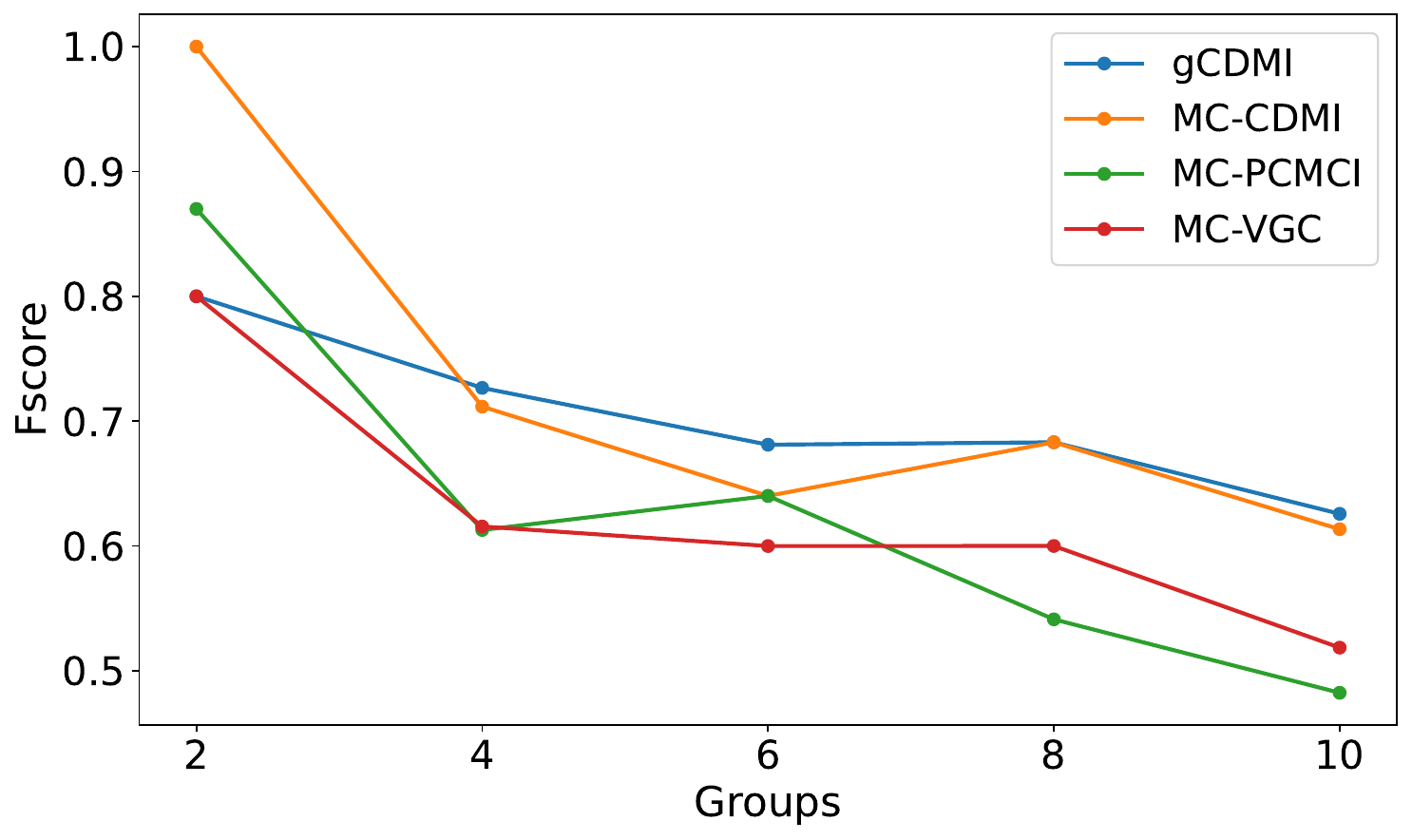}% 
\caption{Performance of group causality methods on the \textit{NetSim} dataset across various network topologies, evaluated using F-scores.} 
\label{fig:multi-netsim}
\end{figure}

Further, we conducted experiments utilizing multi-groups topologies from the \textit{NetSim} datasets, where each group contains 5 variables. Figure \ref{fig:multi-netsim} presents the F-score performance of various causal discovery methods in these scenarios. The gCDMI method consistently achieves the highest F-scores in most cases; however, its performance diminishes as the number of groups increases, similar to synthetic data experiments, suggesting challenges in disentangling complex interactions among brain regions. Conversely, MC-CDMI starts strong but falls behind gCDMI as the group count rises. MC-PCMCI also begins with promising results but experiences significant declines with an increasing number of groups, indicating sensitivity to increasing data dimensionality. While MC-VGC maintains relative stability, it consistently underperforms compared to gCDMI and MC-CDMI, likely due to its limitations in distinguishing causal relationships in intricate multi-group settings; however, towards the end, it yields better results compared to MC-PCMCI. Overall, the methods have difficulty revealing the complex interactions in brain networks compared to other previous datasets.
% --------------------------------------------- 
%                  fMRI-Fscore 
% ---------------------------------------------
% --------------------------------------------
% % ------------------------------------------
% \paragraph{fNIRS data}
% We used fNIRS to find out dyadic interaction by conducting trails between subjects. The subjects initiate events, i.e., hand tapping, foot tapping, and no movements. The other subjects imitate the first subjects. We use these methods to find out how they interact.
% % ------------------------------------------
%                 Discussion
% --------------------------------------------
\section{Discussion}
\label{discussion}

Our experimental results demonstrate that gCDMI generally outperforms other group causality methods across a diverse range of datasets, including synthetic, environmental, geophysical, and neuroimaging data. This improved performance stems from gCDMI’s ability to effectively capture complex nonlinear relationships and interactions at the group level, making it a robust tool for causal discovery in multivariate time series. However, our findings also indicate that no single method consistently dominates across all scenarios. While gCDMI performs well in many cases, there are situations where alternative approaches may yield better results. 

Moreover, causal discovery without integrating domain knowledge remains extremely challenging; such methods primarily support initial investigations rather than providing definitive conclusions. Interpreting causal outcomes requires collaboration with domain experts, who help contextualize results, guide the integration of prior knowledge, and iterate the analysis process. This highlights the importance of combining computational tools with expert insight for robust causal inference in complex systems. 

All experiments were performed on an NVIDIA GeForce RTX 3060 Ti graphics card, providing reliable performance to efficiently handle complex causal analysis tasks across various datasets.

%----------------------------------------
\subsection{Performance Across Varying Data Characteristics}

Our synthetic data experiments demonstrated that gCDMI effectively identifies causal links even as interaction density and nonlinearity increase. While MC-PCMCI and MC-VGC showed modest performance gains in denser graphs, likely due to their increased sensitivity to detecting more connections, gCDMI maintained stable robustness across varying graph densities and conditions. Nonetheless, gCDMI’s performance varied somewhat, indicating sensitivity to certain structural features of the causal graph.

When analyzing real-world datasets, gCDMI demonstrated strong causal inference capabilities across different domains. In climate-ecosystem interactions, gCDMI correctly identified bidirectional causal links in forest ecosystems, reflecting the strong feedback mechanisms inherent in these systems. The superior performance of gCDMI over methods like 2GVecCI and Vanilla-PC highlights its ability to handle complex dependencies in environmental systems.

For tectonic-climate interactions, gCDMI maintained high F-scores across various regimes, reinforcing the established influence of climatic variables and groundwater on tectonic movements. While MC-CDMI exhibited stable performance, its effectiveness declined in regimes characterized by strong nonlinearity. Our findings underscore the importance of regime-based analysis in uncovering causal influences in geophysical systems.

In the ENSO-BCT analysis, gCDMI and 2GVecCI correctly identified the expected causal influence of ENSO on British Columbia’s climate, with gCDMI achieving the highest percentage of true positives. The failure of Vanilla-PC to infer any causal direction suggests limitations in handling long-range climate dependencies. Similarly, in fMRI brain network analysis, gCDMI outperformed other methods in detecting causal interactions, although all methods struggled with the complexity of high-dimensional brain signals.

\subsection{Scalability and Computational Complexity}

Despite its overall improved performance, gCDMI comes with an increased computational cost due to its reliance on deep network-based structure learning. As the number of groups and variables increases, the method experiences a decline in efficiency, particularly in complex datasets like \textit{NetSim} (fMRI). The degradation in performance for larger group structures suggests that while gCDMI is highly effective for moderate-scale problems, additional computational optimizations are needed for large-scale applications.

Comparing scalability, MC-VGC showed the slowest decline in performance with increasing group sizes, making it a viable alternative in high-dimensional settings. MC-PCMCI, while effective in some scenarios, exhibited considerable sensitivity to graph complexity. The trade-off between accuracy and computational efficiency is an important consideration for practical applications, particularly in real-time or large-scale causal inference tasks.

Overall, our study highlights the value of group-based causal methods for uncovering complex dependencies in time series data. While gCDMI shows strong potential, integrating domain expertise into the analysis loop remains essential for accurate interpretation and meaningful insights. Incorporating expert knowledge—whether through collaboration with domain specialists or leveraging tools such as large language models (LLMs) \cite{takayama2024integrating}—can enhance the reliability and impact of causal discoveries. By addressing scalability and deepening expert integration, gCDMI could evolve into a more robust and versatile tool for causal analysis across diverse fields.

% --------------------------------------------

\section{Conclusion}
\label{conclusion}

In this study, we introduced a deep learning-based approach to identify causal interactions within groups of variables or subsystems. By using deep networks for structure learning in an $N$-variate time series system, we perform invariance testing through group-based interventions to uncover causal relationships between groups of variables. Our method can identify bidirectional causal relationships, revealing mutual influences driven by feedback mechanisms in cause-effect systems. While our approach outperforms other causality methods on both synthetic and real-world datasets, it is important to note that the reliance on deep learning comes with a high computational cost. Our findings underscore the significance of data characteristics in causal discovery, emphasizing the need for adaptive methods since no single approach works universally. Nevertheless, gCDMI's strong performance across various domains demonstrates its potential for broader applications in climate science, geophysics, and neuroscience. Future work will aim to improve the scalability of gCDMI and enhance its ability to handle non-stationarity through more generalizable models capable of adapting to regime shifts. Additionally, integrating domain expertise into the causal discovery process remains an important direction to ensure the interpretability and practical relevance of the results.

% ----------------------------------------------
\section*{Acknowledgment}

This work is funded by the Deutsche Forschungsgemeinschaft (DFG, German Research Foundation) –  Individual Research Grant SH 1682/1-1. Maha Shadaydeh is funded by the ERC Synergy Grant: Understanding and Modelling of the Earth System with Machine Learning (USMILE). 
% we need to add USMILE as my fund
% ---------------------------------------------- 
%                   Appendices 
% ----------------------------------------------
\appendices
\label{app:a}
\section{: Knockoff's properties and diagnostics}

\begin{figure*}[t]
\centering
\includegraphics[width=0.8\textwidth]{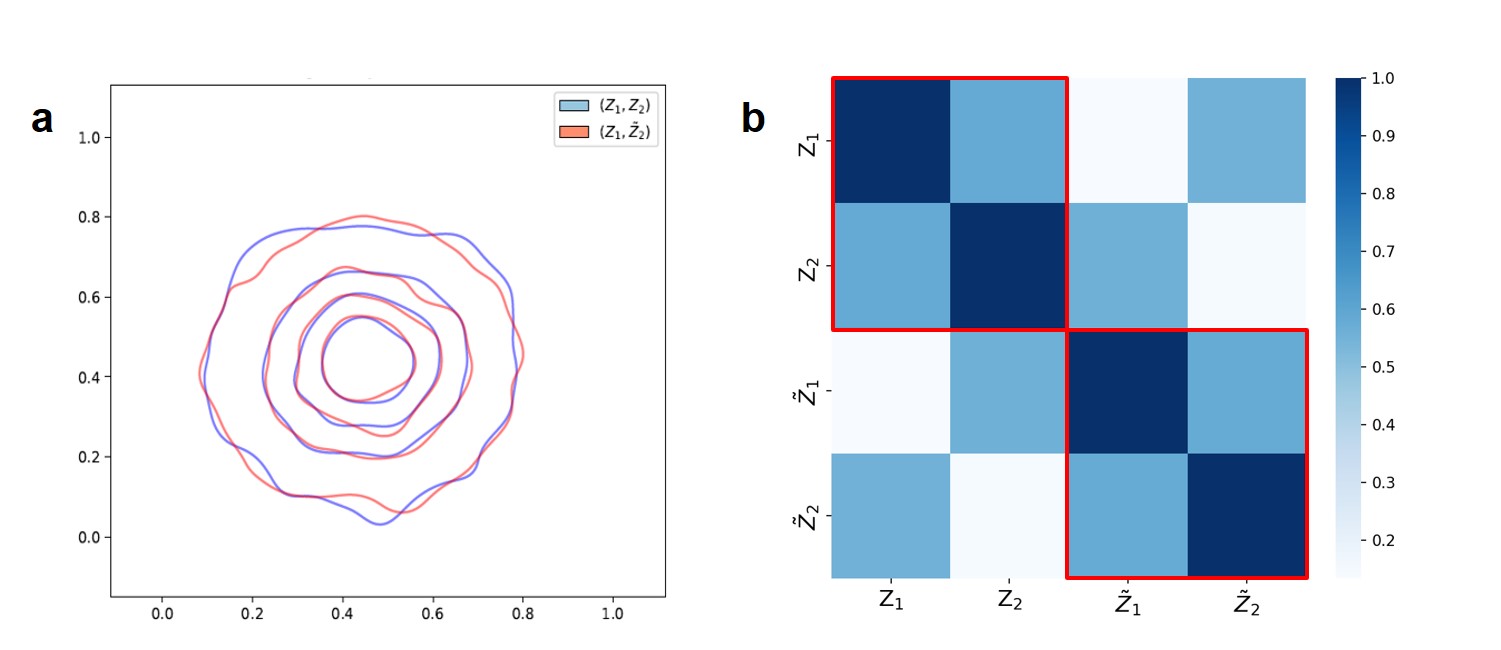}% 
\caption{\textbf{a}. Illustration of exchangeability property \((Z_1, Z_2) \overset{d}{=} (Z_1, \tilde{Z}_2)\) of knockoff variables \textbf{b}. Shows the correlation matrix $\Sigma_{Z\tilde{Z}}$ of the original variables $Z$ and knockoffs $\tilde{Z}$ where the sub-matrix for the correlation within the generated knockoffs \(\Sigma_{\tilde{Z}_{1}\tilde{Z}_{2}}\) is similar to that of correlation in original variables \(\Sigma_{Z_{1}Z_{2}}\), enclosed in red squares. While the variable-wise correlation is minimized with their respective knockoff copies, i.e., \(\sigma_{Z_{1}\tilde{Z}_{1}}, \sigma_{Z_{2}\tilde{Z}_{2}}\approx 0\), shown in white squares. \((\tilde{Z}_1, \dots, \tilde{Z}_N) \bot Y | (Z_1, \dots, Z_N)\)} 
\label{fig:knockoffs}
\end{figure*}

The concept of knockoff variables relies on the exchangeability property, which ensures that substituting a variable with its knockoff counterpart does not alter the underlying joint distribution, i.e., \((Z_1, Z_2) \overset{d}{=} (Z_1, \tilde{Z}_2)\). This property is essential for maintaining the validity of causal inference, particularly in high-dimensional contexts, by enabling robust variable selection and causal discovery while controlling false discoveries. Figure \ref{fig:knockoffs} illustrates the exchangeability property in detail. Panel (a) shows the construction of knockoff variables that satisfy the exchangeability condition. Panel (b) displays the correlation matrix \(\Sigma_{Z\tilde{Z}}\) for the original variables \(Z\) and their corresponding knockoffs \(\tilde{Z}\). As highlighted by the red squares, the sub-matrix representing the correlations among the knockoffs \(\Sigma_{\tilde{Z}_{1}\tilde{Z}_{2}}\) closely mirrors the correlation structure of the original variables \(\Sigma_{Z_{1}Z_{2}}\). This ensures that the knockoffs maintain a similar dependence structure as the original variables, effectively replicating the real-world system. Furthermore, the construction of knockoffs guarantees minimal correlation between each original variable and its corresponding knockoff, i.e., \(\sigma_{Z_{1}\tilde{Z}_{1}}, \sigma_{Z_{2}\tilde{Z}_{2}}\approx 0\), as shown in the white squares of the correlation matrix. This property is crucial for ensuring conditional independence \((\tilde{Z}_1, \dots, \tilde{Z}_N) \bot Y | (Z_1, \dots, Z_N)\), facilitating accurate inference and controlling false discoveries in causal analysis. However, as the number of variables increases, so does the correlation between knockoffs, which makes it more challenging for causal methods to distinguish between true causal relationships and spurious ones.

\begin{figure}
\centering
\includegraphics[width=0.45\textwidth]{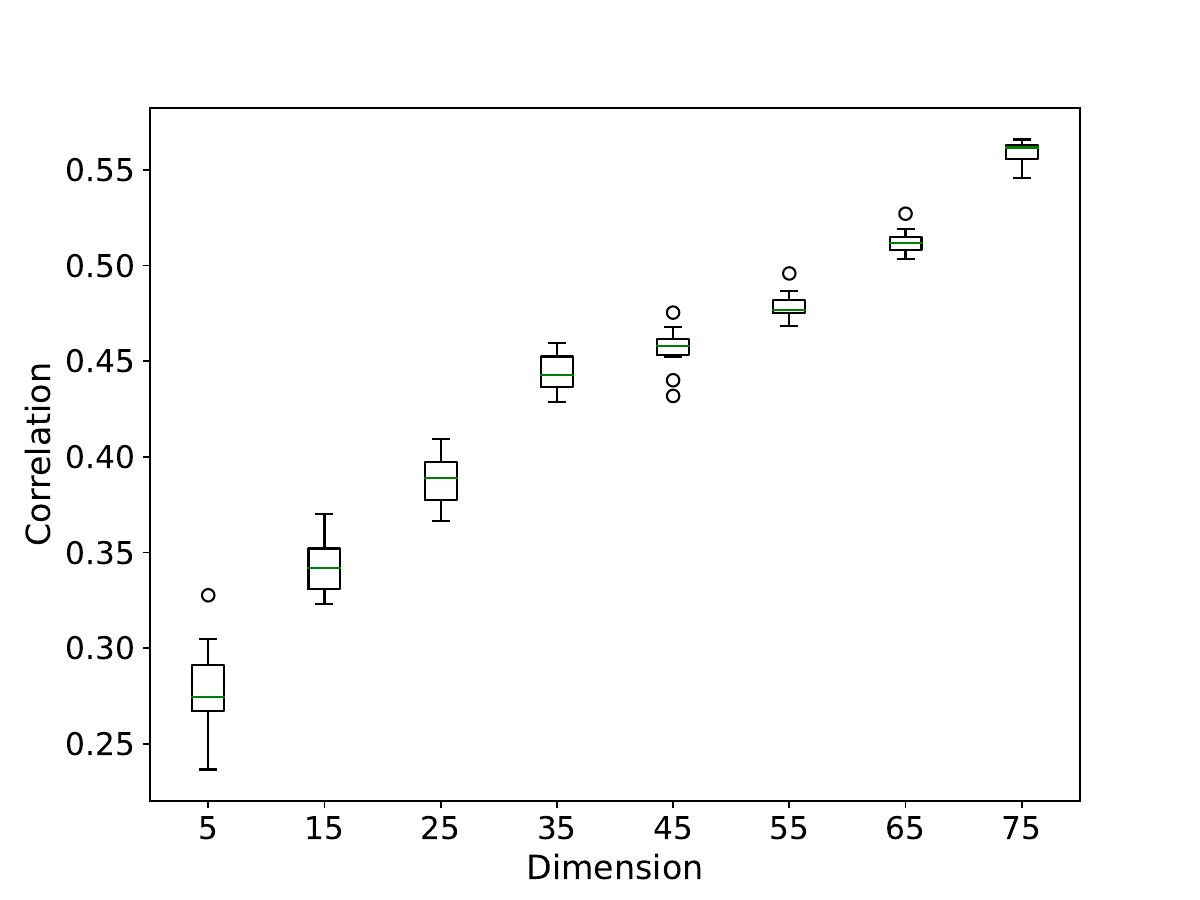}% 
\caption{Pairwise correlations between original variables \((Z_1, \dots, Z_N)\) and their knockoffs \((\tilde{Z}_1, \dots, \tilde{Z}_N)\) across increasing data dimensions. In lower dimensions, the knockoff framework effectively minimizes these correlations, but this property weakens as dimensionality increases.} 
\label{fig:powknockoffs}
\end{figure}

Given that we are conducting causal analysis on high-dimensional, multi-group datasets, it is vital to assess the effectiveness of knockoffs in this context. Our analysis shows that as the dimensionality increases, the pairwise correlation between knockoffs—an essential feature of knockoff-based methods—also rises, as depicted in Figure \ref{fig:powknockoffs}. This growing correlation makes it progressively harder for causal discovery methods to differentiate an original variable from its knockoff counterpart. Consequently, the knockoff-intervention framework faces increased difficulty in identifying true causal relationships. This issue may contribute to the decline in the performance of knockoff-based causal discovery methods as the number of groups increases. The heightened difficulty in differentiation can lead to more false decisions or reduced statistical power, ultimately affecting the reliability of causal inferences in complex, high-dimensional settings.
% ----------------------------------------------------------
% ---------------------------------------------------
% ---------------------------------------------------
%              Knockoffs diagnostics 
% ---------------------------------------------------
% --------------------------------------------------------
%             Table: RegId-Causality
% --------------------------------------------------------
\section{: Regime Identification}
\label{app:b}

\begin{figure}
\centering
\includegraphics[width=0.55\textwidth]{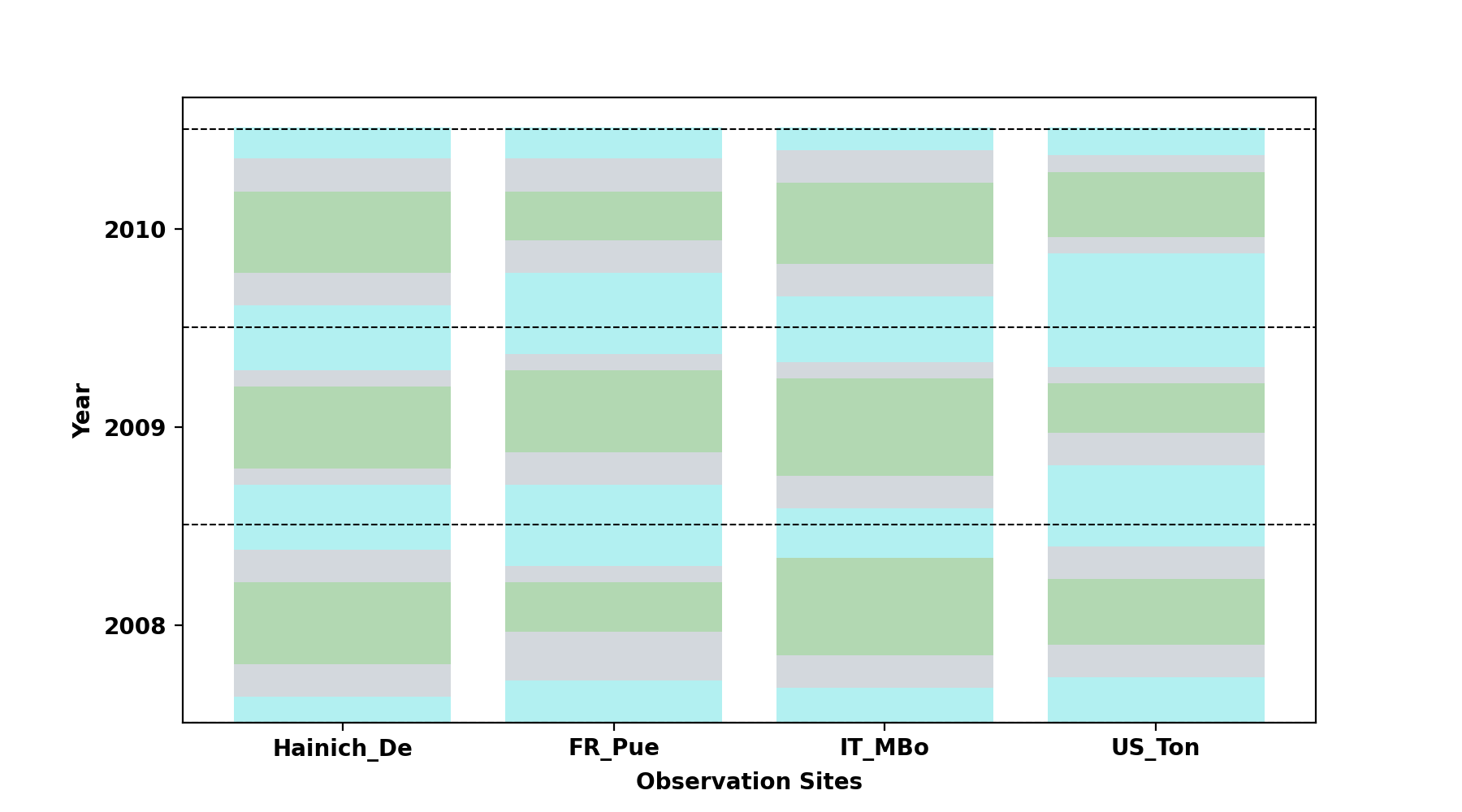}% 
\caption{Identified regimes (blue, gray, green) in various sites in the Fluxnet dataset where summer and winter periods are clearly separated by a transition phase in each year.} 
\label{fig:sitecomp}
\end{figure}

Regime identification enhances causal analysis in non-stationary systems by clustering multivariate time series into distinct regimes based on their covariance matrices \cite{ahmad2024regime}. For the \textit{Fluxnet} dataset, this technique was used to identify periods of local stability by segmenting the data into stable phases, as shown in Figure \ref{fig:sitecomp}. By applying this method, we were able to perform group causal analysis within each identified regime rather than analyzing the entire time series as one continuous dataset. This approach allows for a more nuanced understanding of causal relationships, focusing on stable periods, which is particularly useful when working with non-stationary data.
% ---------------------------------------------------------
    % Regime-wise illustratin of Fluxnet data over sites
% ---------------------------------------------------------
% ---------------------------------------------------------
% ---------------------------------------------------------

% -----------------------------------------------
\section{: Significance testing}
\label{app:c}

In this section, we present properties of some of the statistical hypothesis testing approaches \cite{razali2011power, hazra2016biostatistics, khademi2015statistical} and provide some analysis in regard to causal edge testing in gCDMI.

\textbf{Non-parametric}: The Kolmogorov-Smirnov (KS) test evaluates whether two independent samples come from the same distribution, making no assumptions about the data's underlying distribution. Additionally, the Mann-Whitney U (MWU) test compares the ranks of two independent samples, offering an alternative to the t-test when the data is not normally distributed. The Cramér-von Mises (CM) test, like the KS test, compares the distributions of two independent samples without assuming normality. Lastly, the Wilcoxon Signed-Rank (WSR) test is used for comparing paired samples, similar to the paired t-test, but without the assumption of normality.

\textbf{Parametric}: The Student's t-Test (or Welch’s t-Test) is used to compare the means of two independent samples, under the assumption that the data follows a normal distribution. It helps assess whether there are significant mean differences between the samples. In contrast, the Anderson-Darling (AD) test compares the distributions of two samples, assuming a known distribution, and is primarily used to test the goodness of fit. Additionally, the Shapiro-Wilk (SW) test evaluates the normality of a sample, based on the assumption that the data follows a normal distribution.

\begin{figure*}[ht]
    \centering
    \subfloat[a]{\includegraphics[width=0.5\textwidth]{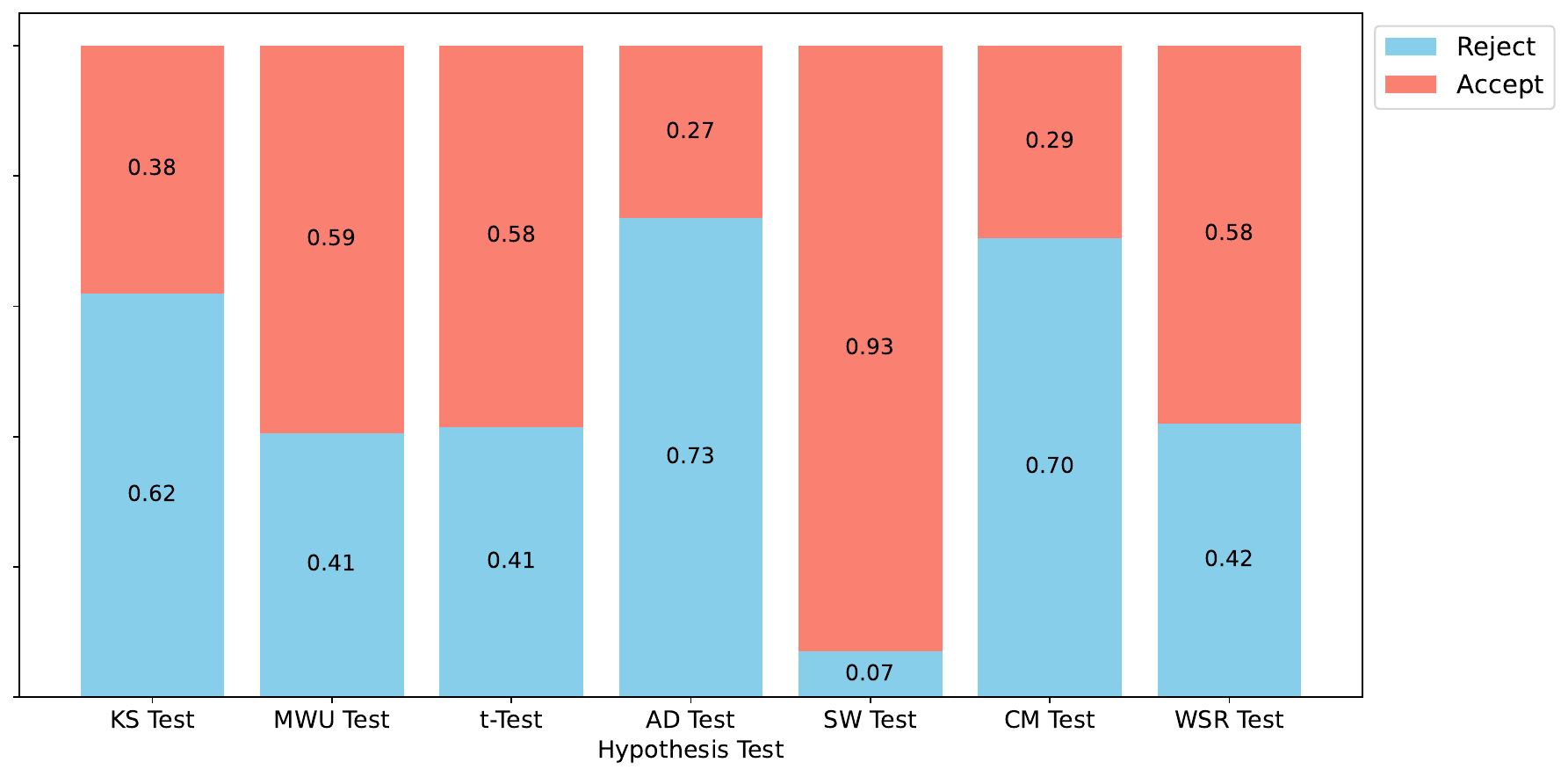}}
     \subfloat[b]{ \includegraphics[width=0.42\textwidth]{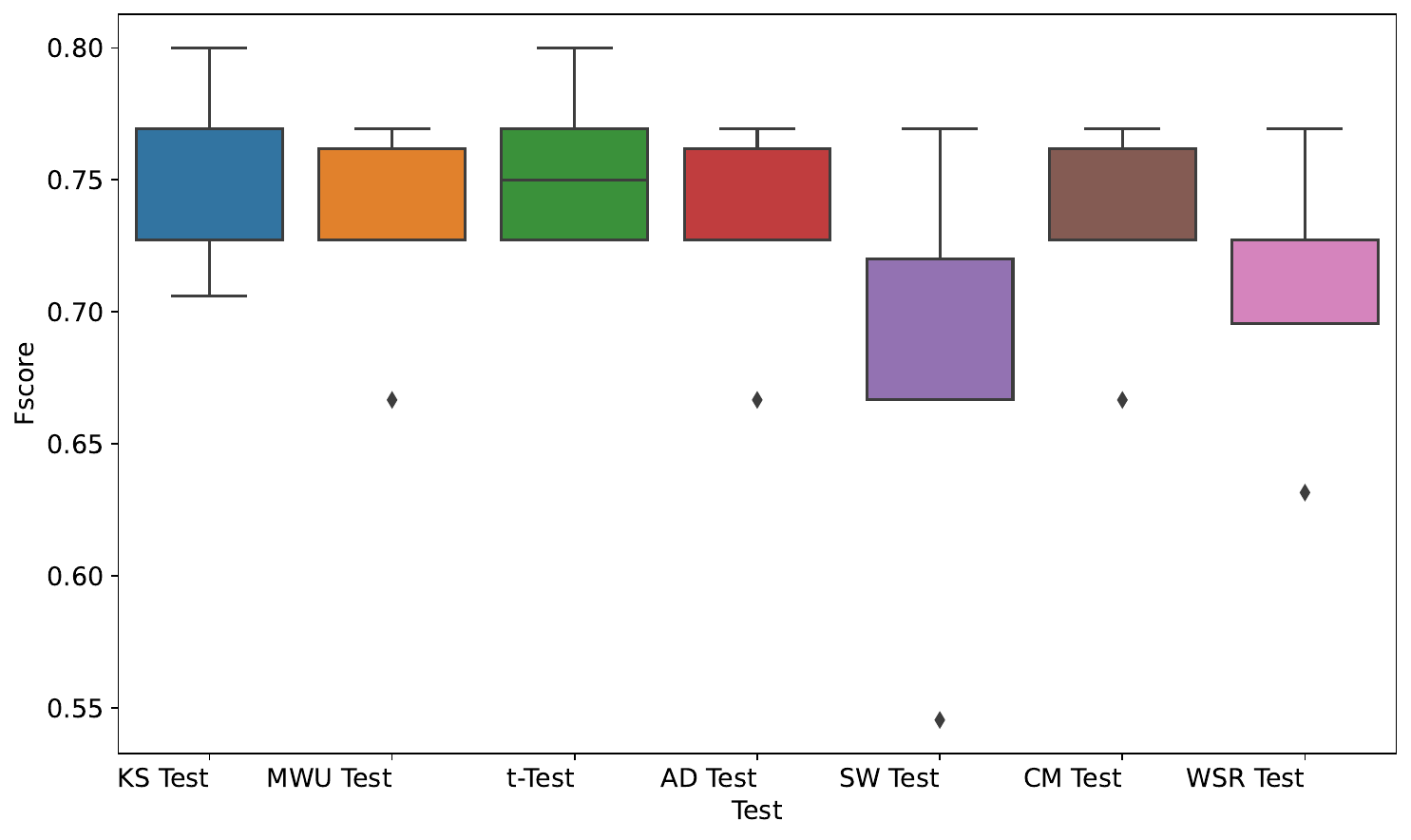}}
    \caption{Significance hypothesis testing (\textbf{a}) Sensitivity of significance testing methods to variations in the statistical properties of the data (mean \(\mu\), variance \(\sigma^2\)) and sample size.  
    (\textbf{b}) F-score results across various statistical hypothesis testing approaches using gCDMI.}
   \label{fig:hype_tests}
\end{figure*}

In Figure \ref{fig:hype_tests} (a), we assessed the sensitivity of various testing methods by manipulating the mean (\(\mu\)), variance (\(\sigma^2\)), both parameters simultaneously, and the number of samples across 200 experiments with two synthetically generated distributions. The results show that the KS-test, Anderson-Darling test, and Cramér-von Mises test successfully detected these variations more than half the time, demonstrating their higher sensitivity to changes. On the other hand, the other tests exhibited a weaker response to these alterations. Following this, we integrated these tests into our causal framework, and in Figure \ref{fig:hype_tests} (b), the alignment between the estimated results and the ground truth causal graphs is shown, highlighting these tests' ability to identify causal relationships. Depending on the analysis's specific context and goals, any of these tests can be selected for optimal performance in detecting causal links.
% -----------------------------------------------------------

% -----------------------------------------------------------
%                       Bibliography
% -----------------------------------------------------------
\bibliographystyle{unsrt}
\bibliography{access}


% \bibitem{b1} G. O. Young, ``Synthetic structure of industrial plastics,'' in \emph{Plastics,} 2\textsuperscript{nd} ed., vol. 3, J. Peters, Ed. New York, NY, USA: McGraw-Hill, 1964, pp. 15--64.

%\vfill

\end{document}

%% file: math_commands.tex
\usepackage{amsmath,amsfonts,bm}

\def\eqref#1{equation~\ref{#1}}
\def\1{\bm{1}}

\DeclareMathAlphabet{\mathsfit}{\encodingdefault}{\sfdefault}{m}{sl}
\SetMathAlphabet{\mathsfit}{bold}{\encodingdefault}{\sfdefault}{bx}{n}